\pdfoutput=1

\documentclass[11pt]{article}
\usepackage{authblk}

\usepackage{acl}
\usepackage{graphicx}
\usepackage{algpseudocode}
\usepackage{algorithm}
\usepackage[switch]{lineno} 
\usepackage{booktabs} 
\usepackage{times}
\usepackage{latexsym}
\usepackage{hyperref}
\usepackage{pifont}
\usepackage{multirow}
\usepackage{amsmath}
\usepackage[T1]{fontenc}
\usepackage[utf8]{inputenc}
\usepackage{microtype}
\usepackage{graphicx}
\usepackage[dvipsnames]{xcolor}
\usepackage{subcaption,siunitx,booktabs}
\usepackage[table]{xcolor}
\usepackage{float}
\DeclareMathOperator*{\argmax}{argmax}

\title{Testing the Assumptions of Active Learning for Translation Tasks with Few Samples}

\author[1,2]{\textbf{Lorenzo Flores}}
\author[1,2]{\textbf{Cesare Spinoso di-Piano}}
\author[1,2]{\textbf{Ori Ernst}}
\author[1,2]{\\\textbf{David Ifeoluwa Adelani}}
\author[1,2]{\textbf{Jackie Chi Kit Cheung}}
\affil[1]{Mila - Quebec AI Institute\quad $^2$McGill University}

\begin{document}
\maketitle
\begin{abstract}

Active learning (AL) is a training paradigm for selecting unlabeled samples for annotation to improve model performance on a test set, which is useful when only a limited number of samples can be annotated. These algorithms often work by optimizing for the informativeness and diversity of the training data to be annotated. Recent work found that AL strategies fail to outperform random sampling on various language generation tasks when using 100-500 samples. To understand AL's poor performance when only using few samples, we investigate whether the core assumptions underlying AL strategies hold. We find that neither the informativeness nor diversity of the training data, which AL strategies optimize for, are correlated with test set performance. Instead, factors like the ordering of the training samples and interactions with pre-training data have a larger impact on performance. This suggests that future AL methods must take these factors into account in order to work with very few samples. Our code is available at \url{https://github.com/ljyflores/active-learning-assumptions.git}.

\end{abstract}

\section{Introduction}

Active learning (AL) is a training paradigm used to select unlabeled data to annotate and train on, given a specific budget \citep{cohn1996activelearningstatisticalmodels}, which is useful when annotation resources are constrained. This has successfully been applied to machine translation (MT) for low-resource scenarios \citep{zeng-etal-2019-empirical, zhao-etal-2020-active, mohiuddin-etal-2022-data, chimoto-bassett-2022-comet}. However, when annotation budgets are very limited (100-500 samples), AL strategies were found to be ineffective \citep{perlitz-etal-2023-active}, meaning that fine-tuning (FT) on data selected by AL fails to yield better test set performance than randomly selected data.

To understand the poor performance of AL in the low-data scenario, we study whether the assumptions that underlie various AL methods still hold when using very little data. We focus our analyses on MT, which is a suitable use case for AL as evidenced by the aforementioned work. The core assumption of many methods is that selecting and training on either more informative or diverse data should lead to better test performance \citep{perlitz-etal-2023-active}. As such, \textbf{informativeness} strategies work by selecting data which the model exhibits high uncertainty on, which is a proxy for how much information the model gains by being fine-tuned on this data. This assumes that choosing the data which provides the most information will yield the largest gains in test performance. In contrast, \textbf{diversity} strategies choose the samples which least resemble the current training data. This assumes that models perform poorly on samples it has not seen in training, and will benefit most from being trained on these novel samples, as these should improve performance on similar samples in the test set. In practice, a combination of both are used.

In this paper, we validate whether selecting and fine-tuning on data that is more informative or diverse, which AL methods aim to do, is actually associated with better test set performance on MT tasks. Then, we explore other factors that affect performance in low-data scenarios, to motivate new methods that take these into account.

We first confirm that AL strategies fail to beat random sampling at these budgets, using the standard AL setup (Section~\ref{section:rq1}). Our remaining analyses do \emph{not} run AL loops. Instead, they test the assumptions that AL strategies rest on: that the informativeness and diversity of a training set are associated with the test performance of a model fine-tuned on it, and that the samples a model performs poorly on are the most useful to train on. Testing these assumptions requires observing many training sets and their resulting performance, including sets that no AL strategy would choose. An AL loop only ever produces the one subset that its acquisition function selects, which cannot show whether the metric predicts performance.

First, we find that for MT tasks, the assumption that training on more informative and diverse data yields better test performance does not hold. By fine-tuning on multiple subsets of data, we find that neither the informativeness nor diversity of the training data is strongly correlated with model performance. We test the implicit assumption made by these AL strategies, that model performance benefits most from training on the samples which the model performs poorly on. We find that AL strategies tend to pick unlabeled samples which the model does not do well on, whereas models actually achieve better test performance using unlabeled samples which the model already performs reasonably well on.

Second, we find that when using very few samples, the ordering of the samples in fine-tuning and the pre-training data have a larger impact on performance than factors like data informativeness or diversity which AL optimizes for. In particular, we find that the ordering of the samples accounts for more of the variance in performance than the content of the samples themselves. We further analyze this qualitatively with a case study on an English-Filipino MT dataset, where we observe that the model is sometimes able to correctly translate words not present in the training data, which suggests that the model is using knowledge from pre-training. Conversely, models are often unable to utilize the correct vocabulary at test time even after seeing them in the training data. Ultimately, these demonstrate that in low data scenarios, other factors may be as, if not more important that the actual training data towards test performance.

In summary, we ask \textbf{RQ1: Is the AL assumption that training on more informative or diverse data yields better performance true in the low data scenario for MT tasks? RQ2: What other factors impact performance in the low data scenario?} Our findings suggest that in scenarios where very little data (100-500 samples) is available, the characteristics of training data optimized for by AL strategies do not meaningfully correlate with performance. This motivates the need for new methods that use other heuristics for the low data regime, or further exploration into how learning occurs with very few samples.

\section{Related Work}

\paragraph{Active Learning} Active learning (AL) is a training paradigm where data is iteratively selected, annotated, and added to the training pool from a set of unlabeled candidates \cite{cohn1996activelearningstatisticalmodels}. AL has been used to efficiently select subsets that achieve better performance than random sampling on image \cite{kirsch2019batchbaldefficientdiversebatch, labonte2022dropout, gal2016dropoutbayesianapproximationrepresenting} and text classification \cite{Zhang_Lease_Wallace_2017, ein-dor-etal-2020-active, prabhu-etal-2019-sampling, siddhant-lipton-2018-deep} tasks. However, \citet{perlitz-etal-2023-active} found that AL strategies did not outperform a random baseline for generation tasks when choosing 100-500 samples. This may hinder the use of AL in machine translation (MT) for low-resource settings, where reducing annotation costs would be most beneficial, as specialized annotation can cost up to \$5 USD/sentence \cite{aidatalabelersDataAnnotation}. We analyze the systematic underperformance, to better understand AL in the very low-resource setting.

\paragraph{Acquisition Functions} Work in AL often focuses on the acquisition function – the strategy for selecting samples. According to \citet{zhang-etal-2022-survey} there are two broad categories: \textbf{Diversity} strategies maximize the diversity of the training examples selected, measured using word-based \cite{zhao-etal-2020-active, zeng-etal-2019-empirical} or embedding-based \cite{sener2018active} metrics. \textbf{Uncertainty/Informativeness} strategies choose samples which the model is most uncertain about and, thus, from which the model is assumed to learn the most information. These use token probability or entropy \cite{zhao-etal-2020-active, mohiuddin-etal-2022-data}, variance in model responses \cite{Gal2017DeepBA, Schmidt2022CombiningDG, LIU2023101444, zeng-etal-2019-empirical}, or predicted quality scores \cite{chimoto-bassett-2022-comet}. In our work, we validate the effectiveness of these strategies in MT and analyze the relationship between these metrics and model performance.

\paragraph{Data Ordering and Curriculum Learning} The order in which training samples are presented is known to affect model performance. Curriculum learning presents samples in a meaningful order, such as increasing difficulty, and can improve training \citep{platanios2019competencebasedcurriculumlearningneural, wan-etal-2020-self}. Even without a curriculum, random data order has been shown to be a substantial source of variance when fine-tuning pre-trained language models \citep{dodge2020finetuningpretrainedlanguagemodels}, and the order of examples in the prompt strongly affects in-context learning performance \citep{lu-etal-2022-fantastically}. Our work shows that when using only few samples, ordering explains most of the variance in performance, so a strategy that selects samples without considering their order optimizes only the smaller share of what determines performance.

\section{Analysis Setup}

We explain the AL framework whose performance we test in the following section, and the acquisition functions used by informativeness and diversity strategies. We provide experimental details for the analyses in the next sections.

\subsection{Active Learning Set-Up}

\paragraph{Algorithm} At each iteration, we choose a subset $\mathcal{S}_i$ from an unlabeled dataset $\mathcal{D}$ using acquisition function $f_\text{aq}$, label it, and fine-tune a model $\theta$ on it, with the goal of maximizing test performance (Algorithm \ref{alg:AL}), using $b=100,500$ samples. This full loop is used only for the AL validation experiment in Section~\ref{section:rq1}; the analyses of the assumptions (Sections~\ref{section:rq1} and~\ref{section:rq2}) fine-tune on fixed subsets, as explained above.

\begin{algorithm}
\caption{Active Learning Framework}\label{alg:AL}
\begin{algorithmic}
\Require 
\State $\mathcal{D} \text{ (Unlabeled Dataset)}, \theta \text{ (Language Model)}$
\State $b \text{ (Budget per Round)}, n \text{ (Num Rounds)}$
\State $f_{\text{aq}} \text{ (Acquisition Function)}$
\For{$i \gets 1 \textbf{ to } n $}
\For{$j \gets 1 \text{ to } |\mathcal{D}|$}
\State $\text{score}_j \gets f_\text{aq}(\mathcal{D}_j, \theta) $
\EndFor
\State $\mathcal{S}_i \gets \text{argmax}_{I \subset \{ 1, \cdots, n \}: |I| = b } \sum_{i \in I} \text{score}_i $
\State $\text{Label } \mathcal{S}_i$
\State $\text{Finetune } \theta \text{ on } \mathcal{S}_i $
\State $\mathcal{D} \gets \mathcal{D} \textbackslash \mathcal{S}_i$
\EndFor
\end{algorithmic}
\end{algorithm}

\paragraph{Acquisition Functions} We select the samples using various informativeness and diversity strategies. We provide the relevant equations in Appendix \ref{Appendix:al_eqs}.

\paragraph{Baselines} At each AL iteration, we randomly sample $b$ samples from the unlabeled data.

\paragraph{Informativeness Strategies} These measure the model's uncertainty in its prediction, quantified by:

\textsc{Mean Token Probability (Mean Prob)} Lowest mean token prob \citep{zablotskaia-etal-2023-uncertainty}

\textsc{Mean Token Entropy (Mean Ent)} Highest mean token entropy \citep{perlitz-etal-2023-active, zhao-etal-2020-active}

\textsc{BALD} Highest BALD score, which aims to measure epistemic uncertainty using the difference between the prediction's entropy and the expected entropy over sampled model parameters \citep{houlsby2011bayesianactivelearningclassification, kirsch2019batchbaldefficientdiversebatch, Gal2017DeepBA}

\textsc{Lexical Similarity (Lex Sim)} Lowest similarity between outputs sampled using dropout \citep{Schmidt2022CombiningDG}, where similarity is measured using METEOR \citep{banarjee2005}

\paragraph{Diversity Strategies} These select the most diverse set of samples, with diversity measured by:

\textsc{DelFy} Highest DelFy, which measures total rarity of the words in a sample, by comparing how frequently the words appear with respect to both the labeled and unlabeled corpora \citep{zhao-etal-2020-active}

\textsc{Core Set} Highest L2 distance between the embedding of a given sample and the embedding of the closest sample in the training set \citep{sener2018active, perlitz-etal-2023-active}

\subsection{Experimental Details}

\paragraph{Models} We test Flan-T5 Base \citep{chung-etal-scale-instruction-2022}, Llama 3.1-8B \citep{grattafiori2024llama3herdmodels}, and Gemma-2-2B (IT) \citep{gemmateam2024gemma2improvingopen}. 

\paragraph{Datasets} We use language pairs from NLLB \cite{nllbteam2022languageleftbehindscaling}: English-Afrikaans (Eng-Afr), English-German (Eng-Ger), and English-Filipino (Eng-Fil) for fine-tuning; we sample 10K sentence pairs for the unlabeled set. We use FLORES Plus \cite{nllb-24} as our test set.

\paragraph{Evaluation} In all analyses, we use the average ChrF+ score \cite{popovic-2017-chrf} over the test set. ChrF+ is a character-level F1 score that has been shown to correlate well with human ratings in translation tasks. We chose it because it does not depend on tokenization and is less sensitive than word-level metrics to the rich morphology and spelling variation common in low-resource languages.

\section{RQ1: Testing core AL assumptions in the low data scenario for MT tasks}
\label{section:rq1}

In this section, we first check whether AL indeed underperforms random sampling in the low-data regime for MT, as this was the motivation for testing the subsequent assumptions. We then describe and test two assumptions made by AL strategies.

\subsection{Validating AL Performance}

We test AL on MT using only $b=100,500$ samples, to check if the findings of \citet{perlitz-etal-2023-active} for AL in the low-data regime apply to MT.

\paragraph{Result} AL strategies fail to outperform random sampling when using $b=100,500$ samples. Table \ref{Table:al_baseline} shows that an AL strategy outperforms random sampling in only 7/54 configurations, and even then, only by margins of <1.99 ChrF points.\footnote{We do not apply multiple testing correction for p-values, because if we did, the results would automatically be non-significant since we only use three seeds. Hence, the reported significance results should be interpreted with caution.} The findings are similar for $b=500$ (Appendix \ref{Appendix:val_study}).

\begin{table*}[!htb]\centering
\resizebox{\textwidth}{!}{
\begin{tabular}{lrrrrrrrrrr}\toprule
&\multicolumn{3}{c}{Flan-T5} &\multicolumn{3}{c}{Llama 3.1} &\multicolumn{3}{c}{Gemma 2} \\
\cmidrule(lr){2-4} \cmidrule(lr){5-7} \cmidrule(lr){8-10}
&Eng-Afr &Eng-Ger &Eng-Fil &Eng-Afr &Eng-Ger &Eng-Fil &Eng-Afr &Eng-Ger &Eng-Fil \\\midrule
BALD &7.23 ± 1.8 &42.72 ± 0.4 &19.8 ± 11.3 & \textsuperscript{*}68.29 ± 0.2 &74.18 ± 0.6 & \textsuperscript{*}64.52 ± 0.2 & \textsuperscript{*}60.04 ± 0.3 &68.58 ± 0.4 &59.13 ± 0.6 \\
Core Set &5.03 ± 2.0 &43.18 ± 0.6 &22.9 ± 6.5 & \textsuperscript{*}68.51 ± 0.4 &74.61 ± 0.2 &63.85 ± 0.2 &58.28 ± 1.2 &68.33 ± 0.7 &59.04 ± 0.3 \\
DelFy &3.47 ± 0.7 &42.10 ± 1.4 &27.3 ± 4.0 &67.91 ± 0.3 & \textsuperscript{*}74.90 ± 0.1 &63.31 ± 0.3 & \textsuperscript{*}59.10 ± 0.4 &68.42 ± 0.3 &58.18 ± 0.6 \\
Lex. Sim &8.38 ± 0.9 &41.14 ± 0.3 &20.3 ± 9.8 &67.54 ± 0.1 &74.69 ± 0.0 & 0 ± 0 &57.01 ± 0.1 &68.33 ± 0.2 &58.05 ± 0.7 \\
Mean Ent &5.82 ± 1.4 &42.81 ± 0.8 &18.7 ± 1.5 &67.08 ± 1.3 &74.53 ± 0.2 &63.21 ± 0.1 &59.50 ± 0.9 &68.27 ± 0.3 &59.13 ± 0.8 \\
Mean Prob &8.37 ± 3.1 &42.24 ± 0.6 &22.6 ± 8.5 &66.95 ± 1.3 &74.44 ± 0.2 &63.26 ± 0.1 & \textsuperscript{*}59.12 ± 0.6 &67.83 ± 0.2 &59.38 ± 0.5 \\
Random &7.28 ± 2.5 &42.46 ± 0.7 &31.2 ± 2.0 &67.74 ± 0.1 &74.56 ± 0.0 &63.70 ± 0.4 &58.05 ± 0.4 &68.20 ± 0.7 &58.32 ± 0.7 \\
\bottomrule
\end{tabular}
}
\caption{Various AL baselines fail to outperform random sampling when using $b=100$ samples (reporting test set ChrF score across three seeds), \textsuperscript{*} indicates significant difference from random (one-way Mann-Whitney, $\alpha=0.05$). The $0 \pm 0$ entry (Lex. Sim., Llama 3.1, Eng-Fil) is a failed run; see text.}
\label{Table:al_baseline}
\end{table*}

\subsection{Testing Association of Data Diversity and Informativeness to Test Performance}

We test the AL assumption that training on more informative/diverse data is associated with better test performance (\textbf{Assumption 1}).

\paragraph{Method} We compute the correlation between the informativeness/diversity of a sample of training data, and the test set performance of the model fine-tuned on it. To do this, we first sample 100 subsets with 100 samples each from the unlabeled set. For each subset, we finetune a model with early stopping, and evaluate on the test set. We use random subsets rather than AL-selected ones so that the metrics vary over a wide range of training sets, which is what is needed to test whether a metric predicts performance.

Then, we measure the informativeness/diversity of the training data using different AL metrics.

For \textbf{informativeness} metrics, we compute metrics by sample, then average over the dataset. We compute average token probability and entropy \cite{zhao-etal-2020-active}, lexical similarity \cite{Schmidt2022CombiningDG}, and BALD score \cite{Gal2017DeepBA} (Appendix \ref{Appendix:al_inf_eqs}). We test other uncertainty quantification metrics developed to estimate epistemic uncertainty for language generation models, namely the weighted average of the average token probabilities of the top $k$ beams \citep{malinin2021uncertaintyestimationautoregressivestructured}, the ratio between the sequence probabilities of the top vs. $k$-th beam \citep{flores-etal-2025-improving}, and the KL divergence between outputs sampled with dropout \citep{lakshminarayanan2017simplescalablepredictiveuncertainty}.

For \textbf{diversity} metrics, we compute (1) DelFy \cite{zhao-etal-2020-active} - a word frequency metric with a penalty for previously seen words, (2) L2 Distance \cite{ni-etal-2022-sentence, sener2018active} - the average L2 distance of training examples from the center\footnote{Computed with the hidden state of the encoder's last layer; Center is the average embedding over the training examples}, and (3) the number of unique vocabulary words in the train set (Appendix \ref{Appendix:al_rep_eqs}).

We then compute the Spearman correlation between the AL metrics and test set performance, to check how associated the informativeness or diversity of the training data is with test performance.

Finally, we study how much of the variance in test set performance is explained by the informativeness or diversity of the training data. We regress the test performance jointly on the metrics above using ordinary least squares and report the $R^2$.

\paragraph{Result} Overall, we find that \textbf{Assumption 1} does not hold in the low data scenario. As shown in Table \ref{Table:performance_vs_al_metric_corr}, the informativeness of the training data is only weakly positively, or even negatively correlated with performance. Diversity metrics for the training data show correlations that are also small (below 0.30 in absolute value), with only one correlation being significantly different from zero ($\alpha=0.05$, Bonferroni correction) across all models and datasets.

\begin{table*}[!htb]\centering
\resizebox{\textwidth}{!}{
\begin{tabular}{lrrrrrrrrrrr}\toprule
& &\multicolumn{3}{c}{Flan-T5} &\multicolumn{3}{c}{Llama 3.1} &\multicolumn{3}{c}{Gemma 2} \\
\cmidrule(lr){3-5} \cmidrule(lr){6-8} \cmidrule(lr){9-11}
& &Eng-Afr &Eng-Ger &Eng-Fil &Eng-Afr &Eng-Ger &Eng-Fil &Eng-Afr &Eng-Ger &Eng-Fil \\
\midrule
\parbox[t]{1mm}{\multirow{4}{*}{\rotatebox[origin=c]{90}{Diversity}}} &Vocabulary Size &-0.0420 &0.0438 &0.0162 &0.0121 &0.0153 &-0.0802 &-0.0131 &-0.0214 &0.117 \\
&DelFy (Source) &0.1256 &-0.0521 &0.0267 &-0.0960 &0.1124 &0.0724 &-0.0145 &0.0086 &-0.2071 \\
&DelFy (Target) & \textsuperscript{*}0.2924 &-0.0819 &0.1646 &-0.0781 &-0.0301 &0.0622 &0.0459 &-0.0479 &0.0721 \\
&L2 Distance &-0.0454 &0.0693 &0.0002 &0.0243 &0.1683 &-0.0300 &-0.0366 &0.0878 &0.1018 \\
\cmidrule{2-11}
\parbox[t]{1mm}{\multirow{7}{*}{\rotatebox[origin=c]{90}{Informativeness}}} &Avg Token Ent &-0.0012 &-0.0589 &-0.1143 &0.0237 &0.0746 &0.0461 &-0.1482 &0.0855 &0.0487 \\
&Avg Token Prob &-0.0234 &0.0756 &0.1751 &-0.0146 &-0.0570 &-0.0680 &0.1312 &-0.0892 &-0.0384 \\
&BS Wt Avg &-0.0373 &0.0489 &0.0296 &-0.0536 &-0.0845 &-0.0435 &0.1462 &-0.0811 &-0.0805 \\
&BS Ratio &-0.0009 &0.0944 &0.1038 &0.2333 &-0.0387 &-0.0812 &0.1267 &0.0168 &-0.0558 \\
&BALD &-0.0041 &0.0104 &0.0702 &0.0363 &-0.0851 &0.0652 &0.0025 &-0.0507 &0.0902 \\
&DO KL Div &0.0073 &-0.0024 &0.0422 &0.0707 &-0.0957 &0.1042 &-0.1063 &-0.0083 &-0.1528 \\
&DO Lexical Sim &0.0089 &-0.0544 &-0.0813 &0.0153 &0.0274 &0.0705 &0.0464 &-0.0201 &-0.0165 \\
\cmidrule{2-11}
& $R^2$ & 0.077 & 0.049 & 0.141 & 0.100 & 0.075 & 0.092 & 0.097 & 0.051 & 0.150 \\
\bottomrule
\end{tabular}
}
\caption{Spearman correlation between AL metrics and model performance with 100 samples for training (Test Set ChrF), * displayed for correlations significantly different from zero ($\alpha=0.05$ with Bonferroni correction)}
\label{Table:performance_vs_al_metric_corr}
\end{table*}

\begin{figure}[htb]
\centering
  \includegraphics[width=\columnwidth]{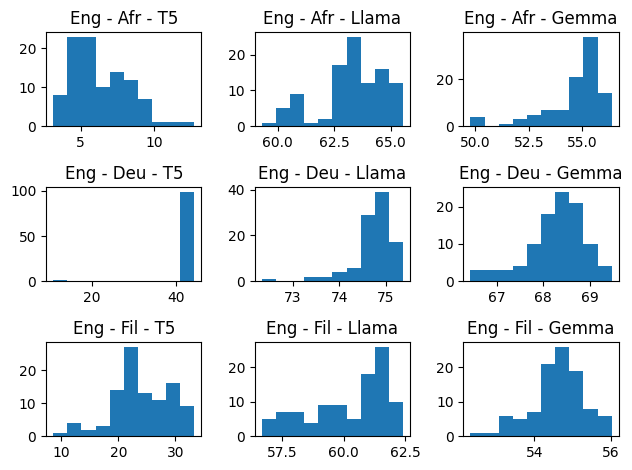}
  \caption{Fine-tuning on different subsets of the data yields considerable variance in test set performance; Plotted using 100 subsets with 100 samples each \label{Figure:sanity_check}}
\end{figure}

While there is wide variance in test performance across subsets (See Figure \ref{Figure:sanity_check}), this is only weakly explained by AL metrics. Informativeness and diversity metrics jointly explain only between 5.1\% to 15.0\% of the total variance in performance when (100 samples), and 2.9 to 19.7\% (500 samples). This suggests that AL metrics only loosely determine performance, and challenges the core assumption of AL that optimizing for these metrics of informativeness or diversity yield better performance.

\subsection{Testing Impact of Fine-Tuning on Samples that Models Perform Poorly On} 

We test the assumption that model performance benefits most from training on samples which the model performs poorly on (\textbf{Assumption 2}).

\paragraph{Method} We first study which samples AL strategies choose; in particular, we study if they choose samples which the models perform well on or poorly on. We take the pre-trained model and generate its predictions for all the samples in the unlabeled set. We then identify which of these samples were chosen by each AL strategy, and plot the model's performance pre-SFT on those samples.

We then study which samples language models benefit most from being trained on. We sort unlabeled candidates by difficulty (measured using the pre-SFT model's performance), then divide them into deciles - the top decile contains samples the pre-SFT model performs very well on, and the lowest decile contains those the model performs worst on. We sample and fine-tune models on 500 unlabeled candidates from each decile.

\begin{figure*}[!htb]
    \centering
    \begin{minipage}{0.48\textwidth}
        \centering
        \includegraphics[width=.9\textwidth]{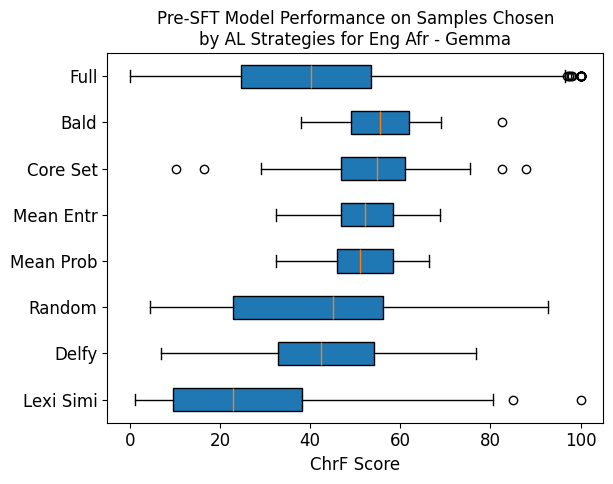}
    \end{minipage}
    \hfill
    \begin{minipage}{0.48\textwidth}
        \centering
        \includegraphics[width=.9\textwidth]{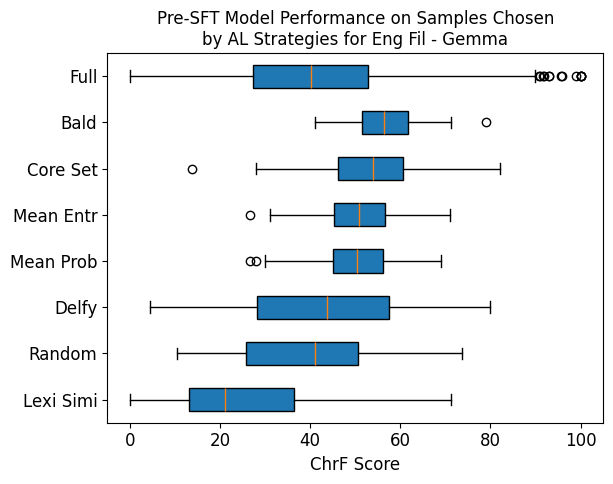}
    \end{minipage}
    \caption{Pre-SFT model performance on the samples chosen by AL strategies for Gemma-2 on Eng-Afr (left) and Eng-Fil (right) shows that AL strategies tend to choose samples in the lower and middle range of pre-SFT performance, rather than samples the model already translates well; strategies vary in where they concentrate}
    \label{fig:performance_probe}
\end{figure*}

\paragraph{Result} First, we observe that AL strategies choose samples which the model performs poorly or mediocrely on, in the sense that they concentrate on the lower and middle of the difficulty range rather than on samples the model already translates well (Figure \ref{fig:performance_probe}). In both plots, the distribution of the entire dataset (Full) spans 0 to 100 ChrF points, but most AL strategies largely pick samples between the 10 to 60 point range. The effect is not uniform: strategies differ in where they concentrate, and the median of some strategies is above the median of the full dataset. Hence, AL strategies tend to avoid the easiest samples, not that every strategy selects uniformly poor-performing ones.

\begin{figure*}[!htb]
\centering
  \includegraphics[width=\textwidth]{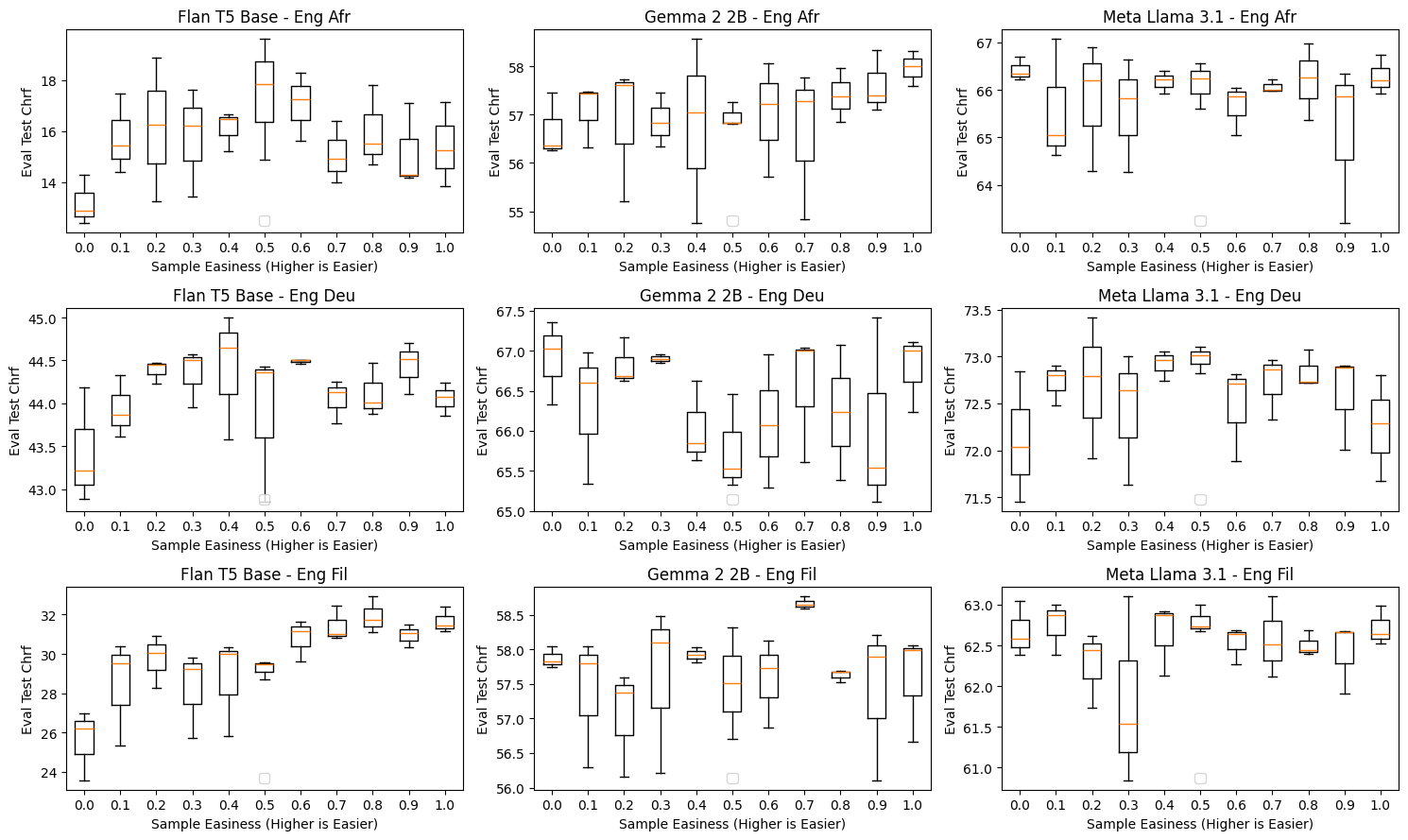}
  \caption{Test set ChrF+ (FLORES Plus) when fine-tuning models on unlabeled data (NLLB) with varying degrees of difficulty (using 500 unlabeled samples, 3 seeds, difficulty measured using pre-SFT model performance) \label{Figure:difficulty_ablation}}
\end{figure*}

We then study which data yields the best test set performance, when fine-tuning models with varying degrees of difficulty. We observe in Figure \ref{Figure:difficulty_ablation} that models achieve better test set performance when being fine-tuned on samples which the model already performed well or mostly well on, with the trends appearing more pronounced when using more samples (Fig \ref{Figure:difficulty_ablation_2000}). This echoes the findings of \citet{swayamdipta2020datasetcartographymappingdiagnosing}, who found that in the classification setting, fine-tuning on data which the model could classify correctly most or all of the time across training epochs yielded the most performance gains. This challenges \textbf{assumption 2}, illustrating the mismatch between what data AL strategies pick and what data actually benefits a model's test set performance.

\paragraph{Possible impact of noisy samples} One reason why models may fail to achieve better performance when trained on the ``difficult'' samples is that these samples may actually be noisy. While we find that this is not the case for the Eng-Fil set (inspected manually), it remains an open question for other languages, and in the usage of AL in general.

\section{RQ2: Impact of FT sample order and pre-training data on performance}
\label{section:rq2}

Since the data's diversity and informativeness did not explain the variance in performance, we turn our attention to other sources of the observed variation. In this section, we identify other factors that strongly impact performance, which may have dwarfed the relationship with diversity and informativeness which AL methods rely on. 

\subsection{Impact of Ordering of FT Samples}

\paragraph{Motivation} We examine sample ordering for two reasons. First, the variance in Figure \ref{Figure:sanity_check} is large even though AL metrics explain little of it, so it must come from something other than the measured properties of the data. Second, prior work on curriculum learning shows that presenting the same data in a different order can change what a model learns \citep{platanios2019competencebasedcurriculumlearningneural, wan-etal-2020-self}, and the variance across random seeds in Table \ref{Table:al_baseline} suggests ordering-like effects even for identical selection strategies.

\paragraph{Method} To compute the proportion of variance in test performance attributed to the ordering of the samples in FT vs. the samples themselves, we decompose the overall variance as follows, where $G$ is the set of sampled subsets, each with $N$ shuffles of the same data, $p_{i,j}$ is the performance (ChrF+ score) from the $i$-th subset with the $j$-th ordering, $\bar{p}_i$ is the average performance for group $i$, and $\bar{p}$ is the average performance across all samples.

\begin{multline*}
    \underbrace{\frac{1}{NG} \sum_{i \in G} \sum_{j = 1}^N (p_{i,j} - \bar{p})^2}_\text{Total Variance} = \\ 
    \underbrace{\frac{1}{NG} \sum_{i \in G} \sum_{j = 1}^N (p_{i,j}-\bar{p_i})^2}_{\substack{\text{Variance within Groups} \\ 
    \text{(from Ordering)}}} + \underbrace{\frac{1}{G} \sum_{i \in G}  (\bar{p_i} - \bar{p})^2}_{\substack{\text{Variance between Groups} \\ \text{(from Sampling)}}}
\end{multline*}

To compute this, we sample $G=10$ subsets of the data, and for each subset, fine-tune models with $N=10$ different seeds, which we verify shuffles the data differently (total 100 models). We use the equation above to compute the proportion of total variance attributed to ordering. Unless otherwise specified, these runs use the default batch size of 8 and learning rate of $5\times10^{-5}$ (Appendix \ref{Appendix:computational}), so the ordering effect we report is measured with mini-batches larger than one.

\paragraph{Result} As shown in Table \ref{Table:variance_decomposition}, we observe a large variance in performance from shuffling the data. In fact, ordering accounts for between 53\% to 90\% of the variance in performance when using 100 samples, and 65\% to 94\% when using 500 samples.

\begin{table}[!htb]
\centering
\begin{tabular}{lrrrr}\toprule
&Flan-T5 &Llama 3.1 &Gemma 2 \\
\midrule
Eng-Afr & 0.81 & 0.71 & 0.80 \\
Eng-Ger & 0.83 & 0.81 & 0.53 \\
Eng-Fil & 0.75 & 0.90 & 0.71 \\
\midrule
Eng-Afr &0.79 &0.92 &0.84 \\
Eng-Ger &0.81 &0.94 &0.65 \\
Eng-Fil &0.93 &0.80 &0.80 \\
\bottomrule
\end{tabular}
\caption{Proportion of variance in ChrF from ordering using 100 (top) and 500 (bottom) samples, computed using ten shuffles of ten subsets of data (100 total)}\label{Table:variance_decomposition}
\end{table}

\subsection{Case Study}

While the results demonstrate the importance of sample order on model performance, it is unclear why some orderings of the data perform better than others. One possibility is that models learn better when samples are presented in increasing difficulty, informativeness, vocabulary diversity, or sample length, as found by previous work in curriculum learning \citep{platanios2019competencebasedcurriculumlearningneural, wan-etal-2020-self}. However, we find no strong evidence of this in our setting. As such, we perform a qualitative analysis across different orderings for one dataset. Our aim is to better understand how the ordering of samples qualitatively affects learning outcomes.

\paragraph{Method} We fine-tune models on multiple shuffles of an English-Filipino task. We chose this translation direction because the authors had access to speakers of Filipino who could verify the outputs; we did not have equivalent access for Afrikaans or German, and a vocabulary-level qualitative analysis is not reliable without it. To isolate the effect of each training sample, this case study uses a batch size of one, unlike the rest of our experiments. At each training step, we analyze how the predictions for the test set change. Because we are studying translation, we focus on the vocabulary learned by the model, which serves as a proxy for the knowledge that the model gains from the training data. The quantitative vocabulary-overlap analysis below (Table \ref{Table:vocab_overlap_corrs}) covers all language pairs and models.

\paragraph{Result} At a high level, our takeaway is that the vocabulary which the model learns is not necessarily the same as the vocabulary in the training data. In particular, the model may fail to use the vocabulary in the training set correctly on the test set, or at all. Moreover, it may generate vocabulary not in the training set, which we hypothesize can only come from the pre-training data. This provides a plausible explanation for why the core assumptions of AL about informativeness and diversity are not met: if models fail to ``learn'' vocabulary correctly, then it does not matter whether you train it on more diverse or informative data, hence AL strategies do not achieve better performance than random. This also suggests that interactions with the model's pre-training data are worth accounting for in future AL methods. Because this analysis covers one translation direction, we treat these findings as hypotheses about why models fail to learn from the data rather than as established results. We find that:

\paragraph{In some orderings, the model learns incorrect translations of the vocabulary} In one shuffle for example, at FT step 91, the model is trained on the word \textit{panalangin} (prayer). After one or more FT steps, the model starts to incorrectly use that word in various test samples. In fact, even after fine-tuning for multiple epochs, the model still incorrectly generates the word \textit{panalangin} in 253 out of 1012 test set examples (See Table \ref{Table:wrong_translation}). This suggests that the model generates the vocabulary en masse without necessarily learning its meaning. In contrast, in another shuffle of the data, the model is fine-tuned on the word \textit{panalangin} at step 14, and does not exhibit this incorrect usage of the word.

\begin{table*}[!htb]
\centering
\scriptsize
\begin{tabular}{p{2.2cm} p{8.8cm} p{3.8cm}}
\toprule
\textbf{Type} & \textbf{Text} & \textbf{Comment} \\
\midrule
\textbf{Source} & ``We now have 4-month-old mice that are non-diabetic that used to be diabetic,'' he added & \\
\midrule
\textbf{Target} & ``Mayroon na tayong 4 na buwang gulang na daga na hindi diabetic na dating diabetic,'' dagdag niya & \\
\midrule
\textbf{Prediction} (Step 23) & ``We now have 4-month-old mice na hindi-diabetic,'' \textbf{katanya}. & Foreign (Indonesian; \textit{katanya}: he said) \\
\midrule
\textbf{Prediction} (Step 70) & ``We ngayon mayroon dalawang buwan \textbf{gulang} na mga maliliit na... katawan ng'' & OOD word (\textit{gulang}: age/old) \\
\bottomrule
\end{tabular}
\caption{\label{Table:param_knowledge_translation} Models generate words not in the training data, both correctly (\textit{gulang}) and incorrectly (\textit{katanya})}

\end{table*}

\begin{figure}[!htb]
\centering
  \includegraphics[width=\columnwidth]{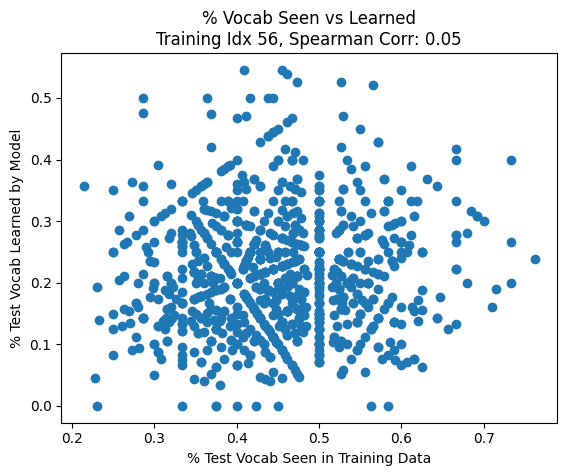}
  \caption{Plot of \% Filipino vocabulary per test example trained on vs. generated at test time (i.e. learned); Training on more vocabulary does not mean the model learns to generate those vocabulary; Each point is generated using a sample in the test set \label{Figure:seen_vs_learned}}
\end{figure}

\paragraph{In some orderings, the models learn less of the vocabulary words in the training data} We see that models are unable to correctly learn certain Filipino vocabulary despite having been trained on them (Figure \ref{Figure:seen_vs_learned}). Moreover, this failure to learn vocabulary is more severe in some shuffles of the data than others. In 4/5 shuffles of the same data, the model fails to generate at least one vocabulary word seen in the training data for 72.1\% of test set examples. However in another shuffle, 85.1\% of test samples have at least one Filipino word which the model does not generate.

\begin{table}[!htb]\centering
\resizebox{\columnwidth}{!}{ 
\begin{tabular}{lrrrr}\toprule
& &\% Covered &\% Similarity \\\midrule
\multirow{3}{*}{Flan-T5} &Eng-Afr &-0.0854 &-0.0994 \\
&Eng-Ger &0.1816 &0.1872 \\
&Eng-Fil &0.1124 &0.1401 \\
\cmidrule{2-4}
\multirow{3}{*}{Llama 3.1} &Eng-Afr &-0.0119 &0.0051 \\
&Eng-Ger &-0.0533 &-0.0788 \\
&Eng-Fil &-0.0727 &-0.0094 \\
\cmidrule{2-4}
\multirow{3}{*}{Gemma 2} &Eng-Afr &0.0291 &0.0249 \\
&Eng-Ger &0.077 &0.0903 \\
&Eng-Fil &0.3266* &0.3676* \\
\bottomrule
\end{tabular}
}
\caption{Spearman correlation between model test set performance and the similarity between the train and test set, using 100 samples for training, * indicates correlation is significantly different from zero ($\alpha=0.05$ with Bonferroni correction)}\label{Table:vocab_overlap_corrs}
\end{table}

We observe this pattern more broadly across the different models and datasets. We take the samples from the previous section, and check whether samples which have more overlapping vocabulary with the test set also achieve higher performance. We measure overlap using the \% of words in the test set present in the training set (\% Covered), and the Jaccard similarity between the training and test set vocabulary (\% Similarity). Intuitively, we expect to see a strong positive correlations between train-test similarity, and test set performance. However, as shown in Table \ref{Table:vocab_overlap_corrs}, the correlation between the two is surprisingly low. Hence, in the low data scenario, even if the training data contains more of the vocabulary used in the test set, it does not necessarily learn to use them correctly at test time.

\paragraph{Some runs also exhibit more interactions with the model's pre-training knowledge} 

We observe that using some shuffles, models generate words not in the training data, which suggests that the model is using data seen in pre-training. Moreover, the extent to which this happens varies by shuffle, which indicates that there are interactions between the ordering of the data and the model's use of its pre-training data.

To illustrate, in one shuffle, the model correctly generates at least one OOD word in 42.7\% of test samples; but only does so for 16.3\% of samples using another shuffle. In Table \ref{Table:param_knowledge_translation}, the model generates \textit{gulang} (age/old), despite it not being in the training corpus. 

Additionally, in some orderings of the data, the model incorrectly generates words from other languages more frequently, despite the training corpus solely being in Filipino. For example, it translates \textit{he added} as \textit{katanya}, which means ``he said'' in Indonesian. This happens across many test set examples\footnote{We identify the languages using Python \texttt{googletrans}}. In some orderings of the data, more test samples have foreign language words (Indonesian: 319, Cebuano: 232), whereas in other orderings, there are fewer (Indonesian: 154, Cebuano: 195 words). It should be noted that these numbers are overestimated as both languages share words with Filipino, but we manually review and confirm that many of them are indeed non-Filipino.

In summary, the case study shows when training on few samples for translation tasks, the relationship between the amount of knowledge (i.e. vocabulary) in the training set and the knowledge the model acquires and uses towards translating the test set is not straightforward. Hence, maximizing for the amount of knowledge in the training set, which AL does by maximizing the number of vocab words covered (diversity strategies), or by choosing samples for which the model is uncertain (informativeness strategies) may not necessarily lead to better performance.

\section{Conclusion}

In this paper, we investigate a core assumption of AL methods that selecting more informative or diverse training data should yield better model performance on the test set. We demonstrate that when little data is available (100 or 500 samples), this assumption does not hold. Moreover, we claim that AL makes an implicit assumption that choosing training data which the model performs poorly on should yield better test performance, and find that the opposite is true. We identify that the impact of the training data's informativeness or diversity may be dwarfed by factors such as sample ordering and interactions with pre-training data, which considerably impact performance as well.

Hence, in low data scenarios, improving model performance is not solely a problem of optimizing for the right informativeness or diversity metrics; it requires understanding the complexities of training and learning involved in translation, and broader generation tasks. Concretely, future work could (1) verify if the results generalize to other generation tasks, (2) analyze and identify interpretable characteristics of the ordering of samples that are associated with better performance to be used as heuristics in future AL algorithms, (3) design AL strategies which select samples that are diverse, informative, and correctly learned by the model, and (4) test whether filtering noisy or misaligned pairs changes how uncertainty-based strategies perform.

\section*{Limitations}

We emphasize that our results are based on very specific model and dataset choices; hence, the current results should not be taken to generalize across all tasks, datasets, and models. We note the following specific limitations.

\paragraph{Evaluation data and metrics} Our evaluation uses a subset of 253 FLORES Plus sentences (25\% of the set) for the main experiments, which limits the precision of the reported scores. We report only ChrF+, chosen for the reasons given in Section 3, and do not report BLEU or learned metrics such as COMET. Conclusions could differ under other metrics, although our correlation and variance analyses depend on relative differences between runs, and the qualitative case study is based on vocabulary rather than on any single score.

\paragraph{Number of seeds and statistical power} The AL comparison in Table \ref{Table:al_baseline} uses three seeds per configuration, and the significance tests there have low power and are uncorrected for multiple comparisons, so they should be read as indicative only. The correlation analyses use 100 subsets, and the ordering analysis uses 10 subsets with 10 shuffles each (100 runs per configuration). We chose these numbers due to the computational cost of the experiments (Appendix \ref{Appendix:computational}); more seeds would give tighter estimates.

\paragraph{Hyperparameters} We test only a specific set of hyperparameters. The main experiments use a batch size of 8 and a constant learning rate of $5\times10^{-5}$ with early stopping, and the case study uses a batch size of one to isolate the effect of individual samples. We did not vary the learning rate or test other batch sizes, and we do not know how large the ordering effect would be under other settings; larger batches average over more samples per update and could reduce it. Even the choice of hyperparameters may yield different model behaviors across runs.

\paragraph{Qualitative analysis} Our section on training dynamics is based off a qualitative study of one translation direction, which the authors chose as they had access to speakers in that language. These results merely serve to provide hypotheses as to why models may fail to learn from the patterns in the data, but more rigorous experimentation is required to make stronger claims about translation or even generation as a whole.

\paragraph{Deployment} We also note that evaluation must be done before deploying any MT model into a real world setting; while AL seeks to improve the performance of these MT models, it should by no means be naively applied and deployed without further testing.

\bibliography{anthology,custom}

\begin{thebibliography}{37}
\expandafter\ifx\csname natexlab\endcsname\relax\def\natexlab#1{#1}\fi

\bibitem[{Banerjee and Lavie(2005)}]{banarjee2005}
Satanjeev Banerjee and Alon Lavie. 2005.
\newblock \href {https://www.aclweb.org/anthology/W05-0909} {{METEOR}: An automatic metric for {MT} evaluation with improved correlation with human judgments}.
\newblock In \emph{Proceedings of the {ACL} Workshop on Intrinsic and Extrinsic Evaluation Measures for Machine Translation and/or Summarization}, pages 65--72, Ann Arbor, Michigan. Association for Computational Linguistics.

\bibitem[{Chimoto and Bassett(2022)}]{chimoto-bassett-2022-comet}
Everlyn~Asiko Chimoto and Bruce~A. Bassett. 2022.
\newblock \href {https://doi.org/10.18653/v1/2022.findings-emnlp.348} {{COMET}-{QE} and active learning for low-resource machine translation}.
\newblock In \emph{Findings of the Association for Computational Linguistics: EMNLP 2022}, pages 4735--4740, Abu Dhabi, United Arab Emirates. Association for Computational Linguistics.

\bibitem[{Chung et~al.(2022)Chung, Hou, Longpre, Zoph, Tay, Fedus, Li, Wang, Dehghani, Brahma, Webson, Gu, Dai, Suzgun, Chen, Chowdhery, Narang, Mishra, Yu, Zhao, Huang, Dai, Yu, Petrov, Chi, Dean, Devlin, Roberts, Zhou, Le, and Wei}]{chung-etal-scale-instruction-2022}
Hyung~Won Chung, Le~Hou, Shayne Longpre, Barret Zoph, Yi~Tay, William Fedus, Eric Li, Xuezhi Wang, Mostafa Dehghani, Siddhartha Brahma, Albert Webson, Shixiang~Shane Gu, Zhuyun Dai, Mirac Suzgun, Xinyun Chen, Aakanksha Chowdhery, Sharan Narang, Gaurav Mishra, Adams Yu, Vincent Zhao, Yanping Huang, Andrew Dai, Hongkun Yu, Slav Petrov, Ed~H. Chi, Jeff Dean, Jacob Devlin, Adam Roberts, Denny Zhou, Quoc~V. Le, and Jason Wei. 2022.
\newblock \href {https://doi.org/10.48550/ARXIV.2210.11416} {Scaling instruction-finetuned language models}.

\bibitem[{Cohn et~al.(1996)Cohn, Ghahramani, and Jordan}]{cohn1996activelearningstatisticalmodels}
D.~A. Cohn, Z.~Ghahramani, and M.~I. Jordan. 1996.
\newblock \href {http://arxiv.org/abs/cs/9603104} {Active learning with statistical models}.

\bibitem[{Dodge et~al.(2020)Dodge, Ilharco, Schwartz, Farhadi, Hajishirzi, and Smith}]{dodge2020finetuningpretrainedlanguagemodels}
Jesse Dodge, Gabriel Ilharco, Roy Schwartz, Ali Farhadi, Hannaneh Hajishirzi, and Noah Smith. 2020.
\newblock \href {http://arxiv.org/abs/2002.06305} {Fine-tuning pretrained language models: Weight initializations, data orders, and early stopping}.

\bibitem[{Ein-Dor et~al.(2020)Ein-Dor, Halfon, Gera, Shnarch, Dankin, Choshen, Danilevsky, Aharonov, Katz, and Slonim}]{ein-dor-etal-2020-active}
Liat Ein-Dor, Alon Halfon, Ariel Gera, Eyal Shnarch, Lena Dankin, Leshem Choshen, Marina Danilevsky, Ranit Aharonov, Yoav Katz, and Noam Slonim. 2020.
\newblock \href {https://doi.org/10.18653/v1/2020.emnlp-main.638} {{A}ctive {L}earning for {BERT}: {A}n {E}mpirical {S}tudy}.
\newblock In \emph{Proceedings of the 2020 Conference on Empirical Methods in Natural Language Processing (EMNLP)}, pages 7949--7962, Online. Association for Computational Linguistics.

\bibitem[{Flores et~al.(2025)Flores, Ernst, and Cheung}]{flores-etal-2025-improving}
Lorenzo Jaime~Yu Flores, Ori Ernst, and Jackie~CK Cheung. 2025.
\newblock \href {https://doi.org/10.18653/v1/2025.acl-short.15} {Improving the calibration of confidence scores in text generation using the output distribution{'}s characteristics}.
\newblock In \emph{Proceedings of the 63rd Annual Meeting of the Association for Computational Linguistics (Volume 2: Short Papers)}, pages 172--182, Vienna, Austria. Association for Computational Linguistics.

\bibitem[{Gal and Ghahramani(2016)}]{gal2016dropoutbayesianapproximationrepresenting}
Yarin Gal and Zoubin Ghahramani. 2016.
\newblock \href {http://arxiv.org/abs/1506.02142} {Dropout as a bayesian approximation: Representing model uncertainty in deep learning}.

\bibitem[{Gal et~al.(2017)Gal, Islam, and Ghahramani}]{Gal2017DeepBA}
Yarin Gal, Riashat Islam, and Zoubin Ghahramani. 2017.
\newblock \href {https://api.semanticscholar.org/CorpusID:6318455} {Deep bayesian active learning with image data}.
\newblock \emph{ArXiv}, abs/1703.02910.

\bibitem[{Grattafiori et~al.(2024)Grattafiori, Dubey, Jauhri, Pandey, Kadian, Al-Dahle, Letman, Mathur, Schelten, Vaughan, Yang, Fan, Goyal, Hartshorn, Yang, Mitra, Sravankumar, Korenev, Hinsvark, Rao, Zhang, Rodriguez, Gregerson, Spataru, Roziere, Biron, Tang, Chern, Caucheteux, Nayak, Bi, Marra, McConnell, Keller, Touret, Wu, Wong, Ferrer, Nikolaidis, Allonsius, Song, Pintz, Livshits, Wyatt, Esiobu, Choudhary, Mahajan, Garcia-Olano, Perino, Hupkes, Lakomkin, AlBadawy, Lobanova, Dinan, Smith, Radenovic, Guzmán, Zhang, Synnaeve, Lee, Anderson, Thattai, Nail, Mialon, Pang, Cucurell, Nguyen, Korevaar, Xu, Touvron, Zarov, Ibarra, Kloumann, Misra, Evtimov, Zhang, Copet, Lee, Geffert, Vranes, Park, Mahadeokar, Shah, van~der Linde, Billock, Hong, Lee, Fu, Chi, Huang, Liu, Wang, Yu, Bitton, Spisak, Park, Rocca, Johnstun, Saxe, Jia, Alwala, Prasad, Upasani, Plawiak, Li, Heafield, Stone, El-Arini, Iyer, Malik, Chiu, Bhalla, Lakhotia, Rantala-Yeary, van~der Maaten, Chen, Tan, Jenkins, Martin, Madaan, Malo, Blecher,
  Landzaat, de~Oliveira, Muzzi, Pasupuleti, Singh, Paluri, Kardas, Tsimpoukelli, Oldham, Rita, Pavlova, Kambadur, Lewis, Si, Singh, Hassan, Goyal, Torabi, Bashlykov, Bogoychev, Chatterji, Zhang, Duchenne, Çelebi, Alrassy, Zhang, Li, Vasic, Weng, Bhargava, Dubal, Krishnan, Koura, Xu, He, Dong, Srinivasan, Ganapathy, Calderer, Cabral, Stojnic, Raileanu, Maheswari, Girdhar, Patel, Sauvestre, Polidoro, Sumbaly, Taylor, Silva, Hou, Wang, Hosseini, Chennabasappa, Singh, Bell, Kim, Edunov, Nie, Narang, Raparthy, Shen, Wan, Bhosale, Zhang, Vandenhende, Batra, Whitman, Sootla, Collot, Gururangan, Borodinsky, Herman, Fowler, Sheasha, Georgiou, Scialom, Speckbacher, Mihaylov, Xiao, Karn, Goswami, Gupta, Ramanathan, Kerkez, Gonguet, Do, Vogeti, Albiero, Petrovic, Chu, Xiong, Fu, Meers, Martinet, Wang, Wang, Tan, Xia, Xie, Jia, Wang, Goldschlag, Gaur, Babaei, Wen, Song, Zhang, Li, Mao, Coudert, Yan, Chen, Papakipos, Singh, Srivastava, Jain, Kelsey, Shajnfeld, Gangidi, Victoria, Goldstand, Menon, Sharma, Boesenberg,
  Baevski, Feinstein, Kallet, Sangani, Teo, Yunus, Lupu, Alvarado, Caples, Gu, Ho, Poulton, Ryan, Ramchandani, Dong, Franco, Goyal, Saraf, Chowdhury, Gabriel, Bharambe, Eisenman, Yazdan, James, Maurer, Leonhardi, Huang, Loyd, Paola, Paranjape, Liu, Wu, Ni, Hancock, Wasti, Spence, Stojkovic, Gamido, Montalvo, Parker, Burton, Mejia, Liu, Wang, Kim, Zhou, Hu, Chu, Cai, Tindal, Feichtenhofer, Gao, Civin, Beaty, Kreymer, Li, Adkins, Xu, Testuggine, David, Parikh, Liskovich, Foss, Wang, Le, Holland, Dowling, Jamil, Montgomery, Presani, Hahn, Wood, Le, Brinkman, Arcaute, Dunbar, Smothers, Sun, Kreuk, Tian, Kokkinos, Ozgenel, Caggioni, Kanayet, Seide, Florez, Schwarz, Badeer, Swee, Halpern, Herman, Sizov, Guangyi, Zhang, Lakshminarayanan, Inan, Shojanazeri, Zou, Wang, Zha, Habeeb, Rudolph, Suk, Aspegren, Goldman, Zhan, Damlaj, Molybog, Tufanov, Leontiadis, Veliche, Gat, Weissman, Geboski, Kohli, Lam, Asher, Gaya, Marcus, Tang, Chan, Zhen, Reizenstein, Teboul, Zhong, Jin, Yang, Cummings, Carvill, Shepard, McPhie,
  Torres, Ginsburg, Wang, Wu, U, Saxena, Khandelwal, Zand, Matosich, Veeraraghavan, Michelena, Li, Jagadeesh, Huang, Chawla, Huang, Chen, Garg, A, Silva, Bell, Zhang, Guo, Yu, Moshkovich, Wehrstedt, Khabsa, Avalani, Bhatt, Mankus, Hasson, Lennie, Reso, Groshev, Naumov, Lathi, Keneally, Liu, Seltzer, Valko, Restrepo, Patel, Vyatskov, Samvelyan, Clark, Macey, Wang, Hermoso, Metanat, Rastegari, Bansal, Santhanam, Parks, White, Bawa, Singhal, Egebo, Usunier, Mehta, Laptev, Dong, Cheng, Chernoguz, Hart, Salpekar, Kalinli, Kent, Parekh, Saab, Balaji, Rittner, Bontrager, Roux, Dollar, Zvyagina, Ratanchandani, Yuvraj, Liang, Alao, Rodriguez, Ayub, Murthy, Nayani, Mitra, Parthasarathy, Li, Hogan, Battey, Wang, Howes, Rinott, Mehta, Siby, Bondu, Datta, Chugh, Hunt, Dhillon, Sidorov, Pan, Mahajan, Verma, Yamamoto, Ramaswamy, Lindsay, Lindsay, Feng, Lin, Zha, Patil, Shankar, Zhang, Zhang, Wang, Agarwal, Sajuyigbe, Chintala, Max, Chen, Kehoe, Satterfield, Govindaprasad, Gupta, Deng, Cho, Virk, Subramanian, Choudhury,
  Goldman, Remez, Glaser, Best, Koehler, Robinson, Li, Zhang, Matthews, Chou, Shaked, Vontimitta, Ajayi, Montanez, Mohan, Kumar, Mangla, Ionescu, Poenaru, Mihailescu, Ivanov, Li, Wang, Jiang, Bouaziz, Constable, Tang, Wu, Wang, Wu, Gao, Kleinman, Chen, Hu, Jia, Qi, Li, Zhang, Zhang, Adi, Nam, Yu, Wang, Zhao, Hao, Qian, Li, He, Rait, DeVito, Rosnbrick, Wen, Yang, Zhao, and Ma}]{grattafiori2024llama3herdmodels}
Aaron Grattafiori, Abhimanyu Dubey, Abhinav Jauhri, Abhinav Pandey, Abhishek Kadian, Ahmad Al-Dahle, Aiesha Letman, Akhil Mathur, Alan Schelten, Alex Vaughan, Amy Yang, Angela Fan, Anirudh Goyal, Anthony Hartshorn, Aobo Yang, Archi Mitra, Archie Sravankumar, Artem Korenev, Arthur Hinsvark, Arun Rao, Aston Zhang, Aurelien Rodriguez, Austen Gregerson, Ava Spataru, Baptiste Roziere, Bethany Biron, Binh Tang, Bobbie Chern, Charlotte Caucheteux, Chaya Nayak, Chloe Bi, Chris Marra, Chris McConnell, Christian Keller, Christophe Touret, Chunyang Wu, Corinne Wong, Cristian~Canton Ferrer, Cyrus Nikolaidis, Damien Allonsius, Daniel Song, Danielle Pintz, Danny Livshits, Danny Wyatt, David Esiobu, Dhruv Choudhary, Dhruv Mahajan, Diego Garcia-Olano, Diego Perino, Dieuwke Hupkes, Egor Lakomkin, Ehab AlBadawy, Elina Lobanova, Emily Dinan, Eric~Michael Smith, Filip Radenovic, Francisco Guzmán, Frank Zhang, Gabriel Synnaeve, Gabrielle Lee, Georgia~Lewis Anderson, Govind Thattai, Graeme Nail, Gregoire Mialon, Guan Pang,
  Guillem Cucurell, Hailey Nguyen, Hannah Korevaar, Hu~Xu, Hugo Touvron, Iliyan Zarov, Imanol~Arrieta Ibarra, Isabel Kloumann, Ishan Misra, Ivan Evtimov, Jack Zhang, Jade Copet, Jaewon Lee, Jan Geffert, Jana Vranes, Jason Park, Jay Mahadeokar, Jeet Shah, Jelmer van~der Linde, Jennifer Billock, Jenny Hong, Jenya Lee, Jeremy Fu, Jianfeng Chi, Jianyu Huang, Jiawen Liu, Jie Wang, Jiecao Yu, Joanna Bitton, Joe Spisak, Jongsoo Park, Joseph Rocca, Joshua Johnstun, Joshua Saxe, Junteng Jia, Kalyan~Vasuden Alwala, Karthik Prasad, Kartikeya Upasani, Kate Plawiak, Ke~Li, Kenneth Heafield, Kevin Stone, Khalid El-Arini, Krithika Iyer, Kshitiz Malik, Kuenley Chiu, Kunal Bhalla, Kushal Lakhotia, Lauren Rantala-Yeary, Laurens van~der Maaten, Lawrence Chen, Liang Tan, Liz Jenkins, Louis Martin, Lovish Madaan, Lubo Malo, Lukas Blecher, Lukas Landzaat, Luke de~Oliveira, Madeline Muzzi, Mahesh Pasupuleti, Mannat Singh, Manohar Paluri, Marcin Kardas, Maria Tsimpoukelli, Mathew Oldham, Mathieu Rita, Maya Pavlova, Melanie Kambadur,
  Mike Lewis, Min Si, Mitesh~Kumar Singh, Mona Hassan, Naman Goyal, Narjes Torabi, Nikolay Bashlykov, Nikolay Bogoychev, Niladri Chatterji, Ning Zhang, Olivier Duchenne, Onur Çelebi, Patrick Alrassy, Pengchuan Zhang, Pengwei Li, Petar Vasic, Peter Weng, Prajjwal Bhargava, Pratik Dubal, Praveen Krishnan, Punit~Singh Koura, Puxin Xu, Qing He, Qingxiao Dong, Ragavan Srinivasan, Raj Ganapathy, Ramon Calderer, Ricardo~Silveira Cabral, Robert Stojnic, Roberta Raileanu, Rohan Maheswari, Rohit Girdhar, Rohit Patel, Romain Sauvestre, Ronnie Polidoro, Roshan Sumbaly, Ross Taylor, Ruan Silva, Rui Hou, Rui Wang, Saghar Hosseini, Sahana Chennabasappa, Sanjay Singh, Sean Bell, Seohyun~Sonia Kim, Sergey Edunov, Shaoliang Nie, Sharan Narang, Sharath Raparthy, Sheng Shen, Shengye Wan, Shruti Bhosale, Shun Zhang, Simon Vandenhende, Soumya Batra, Spencer Whitman, Sten Sootla, Stephane Collot, Suchin Gururangan, Sydney Borodinsky, Tamar Herman, Tara Fowler, Tarek Sheasha, Thomas Georgiou, Thomas Scialom, Tobias Speckbacher,
  Todor Mihaylov, Tong Xiao, Ujjwal Karn, Vedanuj Goswami, Vibhor Gupta, Vignesh Ramanathan, Viktor Kerkez, Vincent Gonguet, Virginie Do, Vish Vogeti, Vítor Albiero, Vladan Petrovic, Weiwei Chu, Wenhan Xiong, Wenyin Fu, Whitney Meers, Xavier Martinet, Xiaodong Wang, Xiaofang Wang, Xiaoqing~Ellen Tan, Xide Xia, Xinfeng Xie, Xuchao Jia, Xuewei Wang, Yaelle Goldschlag, Yashesh Gaur, Yasmine Babaei, Yi~Wen, Yiwen Song, Yuchen Zhang, Yue Li, Yuning Mao, Zacharie~Delpierre Coudert, Zheng Yan, Zhengxing Chen, Zoe Papakipos, Aaditya Singh, Aayushi Srivastava, Abha Jain, Adam Kelsey, Adam Shajnfeld, Adithya Gangidi, Adolfo Victoria, Ahuva Goldstand, Ajay Menon, Ajay Sharma, Alex Boesenberg, Alexei Baevski, Allie Feinstein, Amanda Kallet, Amit Sangani, Amos Teo, Anam Yunus, Andrei Lupu, Andres Alvarado, Andrew Caples, Andrew Gu, Andrew Ho, Andrew Poulton, Andrew Ryan, Ankit Ramchandani, Annie Dong, Annie Franco, Anuj Goyal, Aparajita Saraf, Arkabandhu Chowdhury, Ashley Gabriel, Ashwin Bharambe, Assaf Eisenman, Azadeh
  Yazdan, Beau James, Ben Maurer, Benjamin Leonhardi, Bernie Huang, Beth Loyd, Beto~De Paola, Bhargavi Paranjape, Bing Liu, Bo~Wu, Boyu Ni, Braden Hancock, Bram Wasti, Brandon Spence, Brani Stojkovic, Brian Gamido, Britt Montalvo, Carl Parker, Carly Burton, Catalina Mejia, Ce~Liu, Changhan Wang, Changkyu Kim, Chao Zhou, Chester Hu, Ching-Hsiang Chu, Chris Cai, Chris Tindal, Christoph Feichtenhofer, Cynthia Gao, Damon Civin, Dana Beaty, Daniel Kreymer, Daniel Li, David Adkins, David Xu, Davide Testuggine, Delia David, Devi Parikh, Diana Liskovich, Didem Foss, Dingkang Wang, Duc Le, Dustin Holland, Edward Dowling, Eissa Jamil, Elaine Montgomery, Eleonora Presani, Emily Hahn, Emily Wood, Eric-Tuan Le, Erik Brinkman, Esteban Arcaute, Evan Dunbar, Evan Smothers, Fei Sun, Felix Kreuk, Feng Tian, Filippos Kokkinos, Firat Ozgenel, Francesco Caggioni, Frank Kanayet, Frank Seide, Gabriela~Medina Florez, Gabriella Schwarz, Gada Badeer, Georgia Swee, Gil Halpern, Grant Herman, Grigory Sizov, Guangyi, Zhang, Guna
  Lakshminarayanan, Hakan Inan, Hamid Shojanazeri, Han Zou, Hannah Wang, Hanwen Zha, Haroun Habeeb, Harrison Rudolph, Helen Suk, Henry Aspegren, Hunter Goldman, Hongyuan Zhan, Ibrahim Damlaj, Igor Molybog, Igor Tufanov, Ilias Leontiadis, Irina-Elena Veliche, Itai Gat, Jake Weissman, James Geboski, James Kohli, Janice Lam, Japhet Asher, Jean-Baptiste Gaya, Jeff Marcus, Jeff Tang, Jennifer Chan, Jenny Zhen, Jeremy Reizenstein, Jeremy Teboul, Jessica Zhong, Jian Jin, Jingyi Yang, Joe Cummings, Jon Carvill, Jon Shepard, Jonathan McPhie, Jonathan Torres, Josh Ginsburg, Junjie Wang, Kai Wu, Kam~Hou U, Karan Saxena, Kartikay Khandelwal, Katayoun Zand, Kathy Matosich, Kaushik Veeraraghavan, Kelly Michelena, Keqian Li, Kiran Jagadeesh, Kun Huang, Kunal Chawla, Kyle Huang, Lailin Chen, Lakshya Garg, Lavender A, Leandro Silva, Lee Bell, Lei Zhang, Liangpeng Guo, Licheng Yu, Liron Moshkovich, Luca Wehrstedt, Madian Khabsa, Manav Avalani, Manish Bhatt, Martynas Mankus, Matan Hasson, Matthew Lennie, Matthias Reso, Maxim
  Groshev, Maxim Naumov, Maya Lathi, Meghan Keneally, Miao Liu, Michael~L. Seltzer, Michal Valko, Michelle Restrepo, Mihir Patel, Mik Vyatskov, Mikayel Samvelyan, Mike Clark, Mike Macey, Mike Wang, Miquel~Jubert Hermoso, Mo~Metanat, Mohammad Rastegari, Munish Bansal, Nandhini Santhanam, Natascha Parks, Natasha White, Navyata Bawa, Nayan Singhal, Nick Egebo, Nicolas Usunier, Nikhil Mehta, Nikolay~Pavlovich Laptev, Ning Dong, Norman Cheng, Oleg Chernoguz, Olivia Hart, Omkar Salpekar, Ozlem Kalinli, Parkin Kent, Parth Parekh, Paul Saab, Pavan Balaji, Pedro Rittner, Philip Bontrager, Pierre Roux, Piotr Dollar, Polina Zvyagina, Prashant Ratanchandani, Pritish Yuvraj, Qian Liang, Rachad Alao, Rachel Rodriguez, Rafi Ayub, Raghotham Murthy, Raghu Nayani, Rahul Mitra, Rangaprabhu Parthasarathy, Raymond Li, Rebekkah Hogan, Robin Battey, Rocky Wang, Russ Howes, Ruty Rinott, Sachin Mehta, Sachin Siby, Sai~Jayesh Bondu, Samyak Datta, Sara Chugh, Sara Hunt, Sargun Dhillon, Sasha Sidorov, Satadru Pan, Saurabh Mahajan,
  Saurabh Verma, Seiji Yamamoto, Sharadh Ramaswamy, Shaun Lindsay, Shaun Lindsay, Sheng Feng, Shenghao Lin, Shengxin~Cindy Zha, Shishir Patil, Shiva Shankar, Shuqiang Zhang, Shuqiang Zhang, Sinong Wang, Sneha Agarwal, Soji Sajuyigbe, Soumith Chintala, Stephanie Max, Stephen Chen, Steve Kehoe, Steve Satterfield, Sudarshan Govindaprasad, Sumit Gupta, Summer Deng, Sungmin Cho, Sunny Virk, Suraj Subramanian, Sy~Choudhury, Sydney Goldman, Tal Remez, Tamar Glaser, Tamara Best, Thilo Koehler, Thomas Robinson, Tianhe Li, Tianjun Zhang, Tim Matthews, Timothy Chou, Tzook Shaked, Varun Vontimitta, Victoria Ajayi, Victoria Montanez, Vijai Mohan, Vinay~Satish Kumar, Vishal Mangla, Vlad Ionescu, Vlad Poenaru, Vlad~Tiberiu Mihailescu, Vladimir Ivanov, Wei Li, Wenchen Wang, Wenwen Jiang, Wes Bouaziz, Will Constable, Xiaocheng Tang, Xiaojian Wu, Xiaolan Wang, Xilun Wu, Xinbo Gao, Yaniv Kleinman, Yanjun Chen, Ye~Hu, Ye~Jia, Ye~Qi, Yenda Li, Yilin Zhang, Ying Zhang, Yossi Adi, Youngjin Nam, Yu, Wang, Yu~Zhao, Yuchen Hao, Yundi
  Qian, Yunlu Li, Yuzi He, Zach Rait, Zachary DeVito, Zef Rosnbrick, Zhaoduo Wen, Zhenyu Yang, Zhiwei Zhao, and Zhiyu Ma. 2024.
\newblock \href {http://arxiv.org/abs/2407.21783} {The llama 3 herd of models}.

\bibitem[{Houlsby et~al.(2011)Houlsby, Huszár, Ghahramani, and Lengyel}]{houlsby2011bayesianactivelearningclassification}
Neil Houlsby, Ferenc Huszár, Zoubin Ghahramani, and Máté Lengyel. 2011.
\newblock \href {http://arxiv.org/abs/1112.5745} {Bayesian active learning for classification and preference learning}.

\bibitem[{Kirsch et~al.(2019)Kirsch, van Amersfoort, and Gal}]{kirsch2019batchbaldefficientdiversebatch}
Andreas Kirsch, Joost van Amersfoort, and Yarin Gal. 2019.
\newblock \href {http://arxiv.org/abs/1906.08158} {Batchbald: Efficient and diverse batch acquisition for deep bayesian active learning}.

\bibitem[{LaBonte et~al.(2022)LaBonte, Muthukumar, and Kumar}]{labonte2022dropout}
Tyler LaBonte, Vidya Muthukumar, and Abhishek Kumar. 2022.
\newblock \href {https://openreview.net/forum?id=3OxII8ZB3A} {Dropout disagreement: A recipe for group robustness with fewer annotations}.
\newblock In \emph{NeurIPS 2022 Workshop on Distribution Shifts: Connecting Methods and Applications}.

\bibitem[{Labs(2025)}]{aidatalabelersDataAnnotation}
Elite~Data Labs. 2025.
\newblock {A}{I} {D}ata {A}nnotation {C}osts in 2025: {P}ricing, {I}nsights \& {V}alue --- aidatalabelers.com.
\newblock \url{https://aidatalabelers.com/how-much-do-ai-data-annotation-services-cost-in-2025-the-complete-guide}.
\newblock [Accessed 17-05-2025].

\bibitem[{Lakshminarayanan et~al.(2017)Lakshminarayanan, Pritzel, and Blundell}]{lakshminarayanan2017simplescalablepredictiveuncertainty}
Balaji Lakshminarayanan, Alexander Pritzel, and Charles Blundell. 2017.
\newblock \href {http://arxiv.org/abs/1612.01474} {Simple and scalable predictive uncertainty estimation using deep ensembles}.

\bibitem[{Liu and Yu(2023)}]{LIU2023101444}
Chuanming Liu and Jingqi Yu. 2023.
\newblock \href {https://doi.org/https://doi.org/10.1016/j.csl.2022.101444} {Uncertainty-aware non-autoregressive neural machine translation}.
\newblock \emph{Computer Speech \& Language}, 78:101444.

\bibitem[{Lu et~al.(2022)Lu, Bartolo, Moore, Riedel, and Stenetorp}]{lu-etal-2022-fantastically}
Yao Lu, Max Bartolo, Alastair Moore, Sebastian Riedel, and Pontus Stenetorp. 2022.
\newblock \href {https://doi.org/10.18653/v1/2022.acl-long.556} {Fantastically ordered prompts and where to find them: Overcoming few-shot prompt order sensitivity}.
\newblock In \emph{Proceedings of the 60th Annual Meeting of the Association for Computational Linguistics (Volume 1: Long Papers)}, pages 8086--8098, Dublin, Ireland. Association for Computational Linguistics.

\bibitem[{Malinin and Gales(2021)}]{malinin2021uncertaintyestimationautoregressivestructured}
Andrey Malinin and Mark Gales. 2021.
\newblock \href {http://arxiv.org/abs/2002.07650} {Uncertainty estimation in autoregressive structured prediction}.

\bibitem[{Mohiuddin et~al.(2022)Mohiuddin, Koehn, Chaudhary, Cross, Bhosale, and Joty}]{mohiuddin-etal-2022-data}
Tasnim Mohiuddin, Philipp Koehn, Vishrav Chaudhary, James Cross, Shruti Bhosale, and Shafiq Joty. 2022.
\newblock \href {https://doi.org/10.18653/v1/2022.findings-emnlp.113} {Data selection curriculum for neural machine translation}.
\newblock In \emph{Findings of the Association for Computational Linguistics: EMNLP 2022}, pages 1569--1582, Abu Dhabi, United Arab Emirates. Association for Computational Linguistics.

\bibitem[{Ni et~al.(2022)Ni, Hernandez~Abrego, Constant, Ma, Hall, Cer, and Yang}]{ni-etal-2022-sentence}
Jianmo Ni, Gustavo Hernandez~Abrego, Noah Constant, Ji~Ma, Keith Hall, Daniel Cer, and Yinfei Yang. 2022.
\newblock \href {https://doi.org/10.18653/v1/2022.findings-acl.146} {Sentence-t5: Scalable sentence encoders from pre-trained text-to-text models}.
\newblock In \emph{Findings of the Association for Computational Linguistics: ACL 2022}, pages 1864--1874, Dublin, Ireland. Association for Computational Linguistics.

\bibitem[{{NLLB Team} et~al.(2024){NLLB Team}, Costa-juss{\`a}, Cross, {\c{C}}elebi, Elbayad, Heafield, Heffernan, Kalbassi, Lam, Licht, Maillard, Sun, Wang, Wenzek, Youngblood, Akula, Barrault, Gonzalez, Hansanti, Hoffman, Jarrett, Sadagopan, Rowe, Spruit, Tran, Andrews, Ayan, Bhosale, Edunov, Fan, Gao, Goswami, Guzm{\'a}n, Koehn, Mourachko, Ropers, Saleem, Schwenk, and Wang}]{nllb-24}
{NLLB Team}, Marta~R. Costa-juss{\`a}, James Cross, Onur {\c{C}}elebi, Maha Elbayad, Kenneth Heafield, Kevin Heffernan, Elahe Kalbassi, Janice Lam, Daniel Licht, Jean Maillard, Anna Sun, Skyler Wang, Guillaume Wenzek, Al~Youngblood, Bapi Akula, Loic Barrault, Gabriel~Mejia Gonzalez, Prangthip Hansanti, John Hoffman, Semarley Jarrett, Kaushik~Ram Sadagopan, Dirk Rowe, Shannon Spruit, Chau Tran, Pierre Andrews, Necip~Fazil Ayan, Shruti Bhosale, Sergey Edunov, Angela Fan, Cynthia Gao, Vedanuj Goswami, Francisco Guzm{\'a}n, Philipp Koehn, Alexandre Mourachko, Christophe Ropers, Safiyyah Saleem, Holger Schwenk, and Jeff Wang. 2024.
\newblock \href {https://doi.org/10.1038/s41586-024-07335-x} {Scaling neural machine translation to 200 languages}.
\newblock \emph{Nature}, 630(8018):841--846.

\bibitem[{Perlitz et~al.(2023)Perlitz, Gera, Shmueli-Scheuer, Sheinwald, Slonim, and Ein-Dor}]{perlitz-etal-2023-active}
Yotam Perlitz, Ariel Gera, Michal Shmueli-Scheuer, Dafna Sheinwald, Noam Slonim, and Liat Ein-Dor. 2023.
\newblock \href {https://doi.org/10.18653/v1/2023.emnlp-main.611} {Active learning for natural language generation}.
\newblock In \emph{Proceedings of the 2023 Conference on Empirical Methods in Natural Language Processing}, pages 9862--9877, Singapore. Association for Computational Linguistics.

\bibitem[{Platanios et~al.(2019)Platanios, Stretcu, Neubig, Poczos, and Mitchell}]{platanios2019competencebasedcurriculumlearningneural}
Emmanouil~Antonios Platanios, Otilia Stretcu, Graham Neubig, Barnabas Poczos, and Tom~M. Mitchell. 2019.
\newblock \href {http://arxiv.org/abs/1903.09848} {Competence-based curriculum learning for neural machine translation}.

\bibitem[{Popovi{\'c}(2017)}]{popovic-2017-chrf}
Maja Popovi{\'c}. 2017.
\newblock \href {https://doi.org/10.18653/v1/W17-4770} {chr{F}++: words helping character n-grams}.
\newblock In \emph{Proceedings of the Second Conference on Machine Translation}, pages 612--618, Copenhagen, Denmark. Association for Computational Linguistics.

\bibitem[{Prabhu et~al.(2019)Prabhu, Dognin, and Singh}]{prabhu-etal-2019-sampling}
Ameya Prabhu, Charles Dognin, and Maneesh Singh. 2019.
\newblock \href {https://doi.org/10.18653/v1/D19-1417} {Sampling bias in deep active classification: An empirical study}.
\newblock In \emph{Proceedings of the 2019 Conference on Empirical Methods in Natural Language Processing and the 9th International Joint Conference on Natural Language Processing (EMNLP-IJCNLP)}, pages 4058--4068, Hong Kong, China. Association for Computational Linguistics.

\bibitem[{Schmidt et~al.(2022)Schmidt, Bartezzaghi, Bogojeska, Malossi, and Vu}]{Schmidt2022CombiningDG}
Maximilian Schmidt, A.~Bartezzaghi, Jasmina Bogojeska, Adelmo Cristiano~Innocenza Malossi, and Thang Vu. 2022.
\newblock \href {https://api.semanticscholar.org/CorpusID:254044648} {Combining data generation and active learning for low-resource question answering}.
\newblock In \emph{International Conference on Artificial Neural Networks}.

\bibitem[{Sener and Savarese(2018)}]{sener2018active}
Ozan Sener and Silvio Savarese. 2018.
\newblock \href {https://openreview.net/forum?id=H1aIuk-RW} {Active learning for convolutional neural networks: A core-set approach}.
\newblock In \emph{International Conference on Learning Representations}.

\bibitem[{Siddhant and Lipton(2018)}]{siddhant-lipton-2018-deep}
Aditya Siddhant and Zachary~C. Lipton. 2018.
\newblock \href {https://doi.org/10.18653/v1/D18-1318} {Deep {B}ayesian active learning for natural language processing: Results of a large-scale empirical study}.
\newblock In \emph{Proceedings of the 2018 Conference on Empirical Methods in Natural Language Processing}, pages 2904--2909, Brussels, Belgium. Association for Computational Linguistics.

\bibitem[{Swayamdipta et~al.(2020)Swayamdipta, Schwartz, Lourie, Wang, Hajishirzi, Smith, and Choi}]{swayamdipta2020datasetcartographymappingdiagnosing}
Swabha Swayamdipta, Roy Schwartz, Nicholas Lourie, Yizhong Wang, Hannaneh Hajishirzi, Noah~A. Smith, and Yejin Choi. 2020.
\newblock \href {http://arxiv.org/abs/2009.10795} {Dataset cartography: Mapping and diagnosing datasets with training dynamics}.

\bibitem[{Team et~al.(2024)Team, Riviere, Pathak, Sessa, Hardin, Bhupatiraju, Hussenot, Mesnard, Shahriari, Ramé, Ferret, Liu, Tafti, Friesen, Casbon, Ramos, Kumar, Lan, Jerome, Tsitsulin, Vieillard, Stanczyk, Girgin, Momchev, Hoffman, Thakoor, Grill, Neyshabur, Bachem, Walton, Severyn, Parrish, Ahmad, Hutchison, Abdagic, Carl, Shen, Brock, Coenen, Laforge, Paterson, Bastian, Piot, Wu, Royal, Chen, Kumar, Perry, Welty, Choquette-Choo, Sinopalnikov, Weinberger, Vijaykumar, Rogozińska, Herbison, Bandy, Wang, Noland, Moreira, Senter, Eltyshev, Visin, Rasskin, Wei, Cameron, Martins, Hashemi, Klimczak-Plucińska, Batra, Dhand, Nardini, Mein, Zhou, Svensson, Stanway, Chan, Zhou, Carrasqueira, Iljazi, Becker, Fernandez, van Amersfoort, Gordon, Lipschultz, Newlan, yeong Ji, Mohamed, Badola, Black, Millican, McDonell, Nguyen, Sodhia, Greene, Sjoesund, Usui, Sifre, Heuermann, Lago, McNealus, Soares, Kilpatrick, Dixon, Martins, Reid, Singh, Iverson, Görner, Velloso, Wirth, Davidow, Miller, Rahtz, Watson, Risdal,
  Kazemi, Moynihan, Zhang, Kahng, Park, Rahman, Khatwani, Dao, Bardoliwalla, Devanathan, Dumai, Chauhan, Wahltinez, Botarda, Barnes, Barham, Michel, Jin, Georgiev, Culliton, Kuppala, Comanescu, Merhej, Jana, Rokni, Agarwal, Mullins, Saadat, Carthy, Cogan, Perrin, Arnold, Krause, Dai, Garg, Sheth, Ronstrom, Chan, Jordan, Yu, Eccles, Hennigan, Kocisky, Doshi, Jain, Yadav, Meshram, Dharmadhikari, Barkley, Wei, Ye, Han, Kwon, Xu, Shen, Gong, Wei, Cotruta, Kirk, Rao, Giang, Peran, Warkentin, Collins, Barral, Ghahramani, Hadsell, Sculley, Banks, Dragan, Petrov, Vinyals, Dean, Hassabis, Kavukcuoglu, Farabet, Buchatskaya, Borgeaud, Fiedel, Joulin, Kenealy, Dadashi, and Andreev}]{gemmateam2024gemma2improvingopen}
Gemma Team, Morgane Riviere, Shreya Pathak, Pier~Giuseppe Sessa, Cassidy Hardin, Surya Bhupatiraju, Léonard Hussenot, Thomas Mesnard, Bobak Shahriari, Alexandre Ramé, Johan Ferret, Peter Liu, Pouya Tafti, Abe Friesen, Michelle Casbon, Sabela Ramos, Ravin Kumar, Charline~Le Lan, Sammy Jerome, Anton Tsitsulin, Nino Vieillard, Piotr Stanczyk, Sertan Girgin, Nikola Momchev, Matt Hoffman, Shantanu Thakoor, Jean-Bastien Grill, Behnam Neyshabur, Olivier Bachem, Alanna Walton, Aliaksei Severyn, Alicia Parrish, Aliya Ahmad, Allen Hutchison, Alvin Abdagic, Amanda Carl, Amy Shen, Andy Brock, Andy Coenen, Anthony Laforge, Antonia Paterson, Ben Bastian, Bilal Piot, Bo~Wu, Brandon Royal, Charlie Chen, Chintu Kumar, Chris Perry, Chris Welty, Christopher~A. Choquette-Choo, Danila Sinopalnikov, David Weinberger, Dimple Vijaykumar, Dominika Rogozińska, Dustin Herbison, Elisa Bandy, Emma Wang, Eric Noland, Erica Moreira, Evan Senter, Evgenii Eltyshev, Francesco Visin, Gabriel Rasskin, Gary Wei, Glenn Cameron, Gus Martins,
  Hadi Hashemi, Hanna Klimczak-Plucińska, Harleen Batra, Harsh Dhand, Ivan Nardini, Jacinda Mein, Jack Zhou, James Svensson, Jeff Stanway, Jetha Chan, Jin~Peng Zhou, Joana Carrasqueira, Joana Iljazi, Jocelyn Becker, Joe Fernandez, Joost van Amersfoort, Josh Gordon, Josh Lipschultz, Josh Newlan, Ju~yeong Ji, Kareem Mohamed, Kartikeya Badola, Kat Black, Katie Millican, Keelin McDonell, Kelvin Nguyen, Kiranbir Sodhia, Kish Greene, Lars~Lowe Sjoesund, Lauren Usui, Laurent Sifre, Lena Heuermann, Leticia Lago, Lilly McNealus, Livio~Baldini Soares, Logan Kilpatrick, Lucas Dixon, Luciano Martins, Machel Reid, Manvinder Singh, Mark Iverson, Martin Görner, Mat Velloso, Mateo Wirth, Matt Davidow, Matt Miller, Matthew Rahtz, Matthew Watson, Meg Risdal, Mehran Kazemi, Michael Moynihan, Ming Zhang, Minsuk Kahng, Minwoo Park, Mofi Rahman, Mohit Khatwani, Natalie Dao, Nenshad Bardoliwalla, Nesh Devanathan, Neta Dumai, Nilay Chauhan, Oscar Wahltinez, Pankil Botarda, Parker Barnes, Paul Barham, Paul Michel, Pengchong Jin,
  Petko Georgiev, Phil Culliton, Pradeep Kuppala, Ramona Comanescu, Ramona Merhej, Reena Jana, Reza~Ardeshir Rokni, Rishabh Agarwal, Ryan Mullins, Samaneh Saadat, Sara~Mc Carthy, Sarah Cogan, Sarah Perrin, Sébastien M.~R. Arnold, Sebastian Krause, Shengyang Dai, Shruti Garg, Shruti Sheth, Sue Ronstrom, Susan Chan, Timothy Jordan, Ting Yu, Tom Eccles, Tom Hennigan, Tomas Kocisky, Tulsee Doshi, Vihan Jain, Vikas Yadav, Vilobh Meshram, Vishal Dharmadhikari, Warren Barkley, Wei Wei, Wenming Ye, Woohyun Han, Woosuk Kwon, Xiang Xu, Zhe Shen, Zhitao Gong, Zichuan Wei, Victor Cotruta, Phoebe Kirk, Anand Rao, Minh Giang, Ludovic Peran, Tris Warkentin, Eli Collins, Joelle Barral, Zoubin Ghahramani, Raia Hadsell, D.~Sculley, Jeanine Banks, Anca Dragan, Slav Petrov, Oriol Vinyals, Jeff Dean, Demis Hassabis, Koray Kavukcuoglu, Clement Farabet, Elena Buchatskaya, Sebastian Borgeaud, Noah Fiedel, Armand Joulin, Kathleen Kenealy, Robert Dadashi, and Alek Andreev. 2024.
\newblock \href {http://arxiv.org/abs/2408.00118} {Gemma 2: Improving open language models at a practical size}.

\bibitem[{Team et~al.(2022)Team, Costa-jussà, Cross, Çelebi, Elbayad, Heafield, Heffernan, Kalbassi, Lam, Licht, Maillard, Sun, Wang, Wenzek, Youngblood, Akula, Barrault, Gonzalez, Hansanti, Hoffman, Jarrett, Sadagopan, Rowe, Spruit, Tran, Andrews, Ayan, Bhosale, Edunov, Fan, Gao, Goswami, Guzmán, Koehn, Mourachko, Ropers, Saleem, Schwenk, and Wang}]{nllbteam2022languageleftbehindscaling}
NLLB Team, Marta~R. Costa-jussà, James Cross, Onur Çelebi, Maha Elbayad, Kenneth Heafield, Kevin Heffernan, Elahe Kalbassi, Janice Lam, Daniel Licht, Jean Maillard, Anna Sun, Skyler Wang, Guillaume Wenzek, Al~Youngblood, Bapi Akula, Loic Barrault, Gabriel~Mejia Gonzalez, Prangthip Hansanti, John Hoffman, Semarley Jarrett, Kaushik~Ram Sadagopan, Dirk Rowe, Shannon Spruit, Chau Tran, Pierre Andrews, Necip~Fazil Ayan, Shruti Bhosale, Sergey Edunov, Angela Fan, Cynthia Gao, Vedanuj Goswami, Francisco Guzmán, Philipp Koehn, Alexandre Mourachko, Christophe Ropers, Safiyyah Saleem, Holger Schwenk, and Jeff Wang. 2022.
\newblock \href {http://arxiv.org/abs/2207.04672} {No language left behind: Scaling human-centered machine translation}.

\bibitem[{Wan et~al.(2020)Wan, Yang, Wong, Zhou, Chao, Zhang, and Chen}]{wan-etal-2020-self}
Yu~Wan, Baosong Yang, Derek~F. Wong, Yikai Zhou, Lidia~S. Chao, Haibo Zhang, and Boxing Chen. 2020.
\newblock \href {https://doi.org/10.18653/v1/2020.emnlp-main.80} {Self-paced learning for neural machine translation}.
\newblock In \emph{Proceedings of the 2020 Conference on Empirical Methods in Natural Language Processing (EMNLP)}, pages 1074--1080, Online. Association for Computational Linguistics.

\bibitem[{Zablotskaia et~al.(2023)Zablotskaia, Phan, Maynez, Narayan, Ren, and Liu}]{zablotskaia-etal-2023-uncertainty}
Polina Zablotskaia, Du~Phan, Joshua Maynez, Shashi Narayan, Jie Ren, and Jeremiah Liu. 2023.
\newblock \href {https://doi.org/10.18653/v1/2023.findings-emnlp.197} {On uncertainty calibration and selective generation in probabilistic neural summarization: A benchmark study}.
\newblock In \emph{Findings of the Association for Computational Linguistics: EMNLP 2023}, pages 2980--2992, Singapore. Association for Computational Linguistics.

\bibitem[{Zeng et~al.(2019)Zeng, Garg, Chatterjee, Nallasamy, and Paulik}]{zeng-etal-2019-empirical}
Xiangkai Zeng, Sarthak Garg, Rajen Chatterjee, Udhyakumar Nallasamy, and Matthias Paulik. 2019.
\newblock \href {https://doi.org/10.18653/v1/D19-6110} {Empirical evaluation of active learning techniques for neural {MT}}.
\newblock In \emph{Proceedings of the 2nd Workshop on Deep Learning Approaches for Low-Resource NLP (DeepLo 2019)}, pages 84--93, Hong Kong, China. Association for Computational Linguistics.

\bibitem[{Zhang et~al.(2017)Zhang, Lease, and Wallace}]{Zhang_Lease_Wallace_2017}
Ye~Zhang, Matthew Lease, and Byron Wallace. 2017.
\newblock \href {https://doi.org/10.1609/aaai.v31i1.10962} {Active discriminative text representation learning}.
\newblock \emph{Proceedings of the AAAI Conference on Artificial Intelligence}, 31(1).

\bibitem[{Zhang et~al.(2022)Zhang, Strubell, and Hovy}]{zhang-etal-2022-survey}
Zhisong Zhang, Emma Strubell, and Eduard Hovy. 2022.
\newblock \href {https://doi.org/10.18653/v1/2022.emnlp-main.414} {A survey of active learning for natural language processing}.
\newblock In \emph{Proceedings of the 2022 Conference on Empirical Methods in Natural Language Processing}, pages 6166--6190, Abu Dhabi, United Arab Emirates. Association for Computational Linguistics.

\bibitem[{Zhao et~al.(2020)Zhao, Zhang, Zhou, and Zhang}]{zhao-etal-2020-active}
Yuekai Zhao, Haoran Zhang, Shuchang Zhou, and Zhihua Zhang. 2020.
\newblock \href {https://doi.org/10.18653/v1/2020.findings-emnlp.162} {Active learning approaches to enhancing neural machine translation}.
\newblock In \emph{Findings of the Association for Computational Linguistics: EMNLP 2020}, pages 1796--1806, Online. Association for Computational Linguistics.

\end{thebibliography}
\bibliographystyle{acl_natbib}

\appendix

\section{AL Metrics}
\label{Appendix:al_eqs}

\subsection{Diversity Metrics}
\label{Appendix:al_rep_eqs}

\paragraph{Delfy \cite{zhang-etal-2022-survey}}
\begin{equation}
\begin{split}
    \label{Eq:Delfy}
    & \text{f}_\text{Delfy}(\mathcal{S}) =\frac{1}{|\mathcal{S}|} \sum_{x \in \mathcal{S}} \text{Delfy}(x) \\
    &\text{Delfy}(x) = \\ 
    & \frac{1}{|x|} \sum_{i=1}^{|x|} \frac{\text{log}(C(x_i|U)+1)}{\sum_{w^\prime \in U} \text{log}(C(w^\prime|U)+1) } \cdot p_\text{Delfy}(x_i) \\
    & \text{lf}(x) = \\ 
    & \frac{1}{|x|} \sum_{i=1}^{|x|} \frac{\text{log}(C(x_i|U)+1)}{\sum_{w^\prime \in U} \text{log}(C(w^\prime|U)+1) } \cdot p_\text{Lf}(x_i) \\
    & p_\text{Delfy}(x_i) = e^{-\lambda_1 C(x_i|L)} \cdot e^{-\lambda_2 C(x_i|\hat{U}(x))} \\
    & p_\text{Lf}(x_i) = e^{-\lambda_1 C(x_i|L)}
    \end{split}
\end{equation}

Where $U$ is the set of untranslated target sentences, $\hat{U}(x)$ is the set of untranslated sentences with $ls$ score higher than $ls(x)$, $L=\{\}$ is the (empty) set of already selected sentences, $C(w|S)$ is the number of times word $w$ appears in a set $S$, and $p_\text{Delfy}$ and $p_\text{Lf}$ are penalty functions to penalize seen words, in which we use $\lambda_2=1$

\paragraph{L2 Distance}
\begin{equation}
    \label{Eq:L2_Dist}
    \text{f}_\text{L2}(\mathcal{S}) = \frac{1}{|\mathcal{S}|} \sum_{x \in \mathcal{S}} ||h_{f_\theta}(x) - \bar{h}_{f_\theta}(\mathcal{S}) ||_2^2
\end{equation}

Where $h_{f_\theta}(x) \in \mathcal{R}^d$ is the hidden state representation of $x$, obtained by taking the last hidden state of encoder $f_\theta$ and averaging it over the vocab, so that it is a vector of dimension $d$, and $\bar{h}_{f_\theta}(\mathcal{S})$ is the average hidden state across all samples in $\mathcal{S}$.

\paragraph{Greedy Core Set \cite{sener2018active}}

We describe one round of the greedy core set by \citet{sener2018active} in Algorithm \ref{alg:greedy_core_set}, where $\Delta(x,y)=|| h_{f_\theta}(x) - h_{f_\theta}(y) ||_2^2$

\begin{algorithm}[h]
\caption{Core Set Algorithm (1 Round)}\label{alg:greedy_core_set}
\begin{algorithmic}
\Require 
\State $\mathcal{D} \text{ (Unlabeled Dataset)}$, $\mathcal{L} \text{ (Labeled Dataset)}$
\State $b \text{ (Budget per Round)}$, $f_{\theta} \text{ (LM)}$
\For{$i \gets 1 \text{ to } b$}
\State $u \gets \argmax_{x \in \mathcal{D}} \min_{y \in \mathcal{L}} \Delta(x,y)$
\State $\mathcal{L} \gets \mathcal{L} \bigcup \{u\}$
\State $\mathcal{D} \gets \mathcal{D} \textbackslash \{u\}$
\EndFor
\end{algorithmic}
\end{algorithm}

\subsection{Informativeness Metrics}
\label{Appendix:al_inf_eqs}

\paragraph{Average Token Probability \& Entropy \cite{zhao-etal-2020-active}}
\begin{equation}
    \label{Eq:Token_Probability}
    \text{f}_\text{ATP}(x) = \frac{1}{|\hat{y}|} \sum_{t=1}^{|\hat{y}|} p(\hat{y}_t | \hat{y}_{<t}, x)
\end{equation}

\begin{equation}
    \label{Eq:Token_Entropy}
    \text{f}_\text{ATE}(x) = \frac{1}{|\hat{y}|} \sum_{t=1}^{|\hat{y}|} \mathcal{H} \left(p(\hat{y}_t | \hat{y}_{<t}, x)\right)
\end{equation}

\paragraph{Lexical Similarity \cite{Schmidt2022CombiningDG}}

\begin{equation}
    \label{Eq:METEOR_Var}
    \text{f}_\text{LS}(x) = \frac{\sum_{i=1}^{10} \sum_{j=1}^{10} \text{Meteor}(\hat{y}^{(i)}, \hat{y}^{(j)})}{N(N-1)} 
\end{equation}

We compute lexical similarity, where similarity is measured using METEOR \cite{banarjee2005}.

\paragraph{BALD \cite{Gal2017DeepBA}}
\begin{equation}
\label{Eq:BALD_Var}
\begin{split}
    \text{f}_\text{BALD}(x) = \frac{1}{|\hat{y}|} \sum_{t=1}^{|\hat{y}|} \mathcal{H} \left(p(\hat{y}_t | \hat{y}_{<t}, x)\right) \\ - \frac{1}{k} \sum_{i=1}^k \frac{1}{|\hat{y}^{(i)}|} \sum_{t=1}^{|\hat{y}^{(i)}|} \mathcal{H} \left(p(\hat{y}^{(i)}_t | \hat{y}^{(i)}_{<t}, x)\right)
\end{split}
\end{equation}

\begin{multline*}
    \mathcal{H}\left(p(\hat{y}_t | \hat{y}_{<t}, x)\right)= \\ - \sum_{j=1}^{|\mathcal{V}|} p(\hat{y}_{t,j} | \hat{y}_{<t}, x) \text{log} \left( p(\hat{y}_{t,j} | \hat{y}_{<t}, x)\right)
\end{multline*}

Where $\hat{y}$ is the predicted output, $\hat{y}^{(i)}$ is the $i$-th predicted output generated by sampling using dropout, and $\hat{y}_t$ and $\hat{y}^{(i)}_t$ are their $t$-th tokens

\section{Fine-Tuning Details}

We run all our experiments on RTX 8000 GPUs; each active learning run in the validation experiment took roughly 10 GPU hours, whereas the sampling and ordering GPU hours took roughly 72 GPU hours per translation direction.

\section{Dataset Details}
We use the NLLB dataset \cite{nllb-24} under the ODC-By License, and the FLORES Plus dataset \cite{nllbteam2022languageleftbehindscaling} under the CC BY-SA 4.0 License, which allow the use of these datasets for research purposes. We scan the datasets to check that there are no malicious or harmful content in the translation pairs. We did not otherwise verify the alignment or quality of the sentence pairs (see Limitations). For these datasets, we use the English-Afrikaans, English-German, and English-Filipino datasets.

For each of the datasets, we sample 100 sentence-pairs from NLLB to use as the initial labeled candidates, and another 10000 pairs as the unlabeled candidates. Then, we use a sample of 253 candidates (25\%) of the FLORES dataset for evaluation.

\section{Computational Details}
\label{Appendix:computational}

Unless otherwise specified, we use a batch size of 8 and constant learning rate of 5e-5. The case study (Section \ref{section:rq2}) is the exception and uses a batch size of one. We train models for a maximum of 200 epochs, but employ early stopping with a patience of 2 epochs; training is stopped once ChrF+ on a held-out validation set of 100 samples, sampled separately from the unlabeled set, degrades. We perform all fine-tuning and inference using one RTX 8000 GPU.

\section{Validation Study Results}
\label{Appendix:val_study}
\begin{table*}[!htp]\centering
\resizebox{\textwidth}{!}{ 
\begin{tabular}{lrrrrrrrrrr}\toprule
&\multicolumn{3}{c}{Flan-T5} &\multicolumn{3}{c}{Llama-3.1} &\multicolumn{3}{c}{Gemma-2} \\
\cmidrule(lr){2-4} \cmidrule(lr){5-7} \cmidrule(lr){8-10}
&Eng-Afr &Eng-Ger &Eng-Fil &Eng-Afr &Eng-Ger &Eng-Fil &Eng-Afr &Eng-Ger &Eng-Fil \\\midrule
BALD &29.58 ± 1.4 &43.88 ± 0.2 &29.7 ± 3.3 &68.91 ± 0.3 &74.53 ± 0.2 &63.43 ± 0.4 &59.62 ± 1.1 &68.38 ± 0.3 &60.29 ± 0.4 \\
Core Set &14.11 ± 4.3 &44.39 ± 0.4 &30.73 ± 3 &68.35 ± 0.6 &74.56 ± 0.3 &63.92 ± 1 &59.16 ± 1.2 &68.17 ± 0.3 &60.32 ± 0.3 \\
DelFy &20.12 ± 1.7 &43.51 ± 0.8 &28.7 ± 0.8 &68.64 ± 0.3 &74.48 ± 0.1 &63.93 ± 0.8 &58.63 ± 1.2 &68.49 ± 0.3 &59.12 ± 0.8 \\
Lex. Sim &11.15 ± 2.9 &43.63 ± 0.2 &28.05 ± 1 &66.57 ± 0.3 &73.72 ± 0.2 &0 ± 0 &55.2 ± 0.5 &68.02 ± 0.4 &57.13 ± 0.8 \\
Mean Ent &23.62 ± 2.8 &43.12 ± 0.7 &29.64 ± 1 &64.95 ± 1.3 &74.24 ± 0.4 &62.97 ± 0.5 &60.36 ± 0.7 &68.3 ± 0.3 &60.15 ± 0.6 \\
Mean Prob &26.02 ± 7.6 &43.26 ± 0.4 &32.14 ± 1 &67.41 ± 0.8 &73.52 ± 0.5 &63.01 ± 1 &59.54 ± 1 &68.15 ± 0.5 &60.67 ± 0.5 \\
Random &28.38 ± 0.3 &43.67 ± 0.5 &31.25 ± 1.1 &68.41 ± 0.6 &74.17 ± 0.3 &64.59 ± 0.2 &59.09 ± 0.6 &68.27 ± 0.5 &58.37 ± 0.9 \\
\bottomrule
\end{tabular}
}
\caption{AL baseline performance with 500 samples with st. deviation reported across 3 seeds (Test Set ChrF). The $0 \pm 0$ entry (Lex. Sim., Llama-3.1, Eng-Fil) is a failed run; see Section \ref{section:rq1}.}\label{tab:}
\end{table*}

\section{Additional Results}
\label{Appendix:additional_results}
\begin{figure}[htb]
\centering
  \includegraphics[width=\columnwidth]{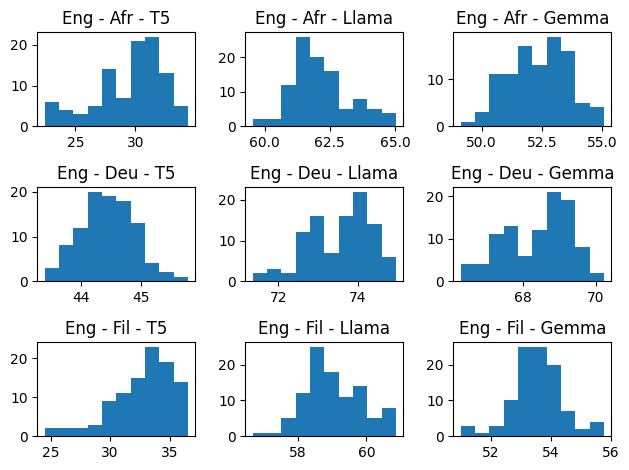}
  \caption{Fine-tuning on different subsets of the data yields considerable variance in test set performance; Plotted using 100 subsets with 500 samples each \label{Figure:sanity_check_500}}
\end{figure}

\begin{table*}[!htp]\centering
\resizebox{\textwidth}{!}{ 
\begin{tabular}{lrrrrrrrrrrr}\toprule
& &\multicolumn{3}{c}{Flan-T5} &\multicolumn{3}{c}{Llama 3.1} &\multicolumn{3}{c}{Gemma 2} \\
\cmidrule(lr){3-5} \cmidrule(lr){6-8} \cmidrule(lr){9-11}
& &Eng-Afr &Eng-Ger &Eng-Fil &Eng-Afr &Eng-Ger &Eng-Fil &Eng-Afr &Eng-Ger &Eng-Fil \\\midrule
\parbox[t]{1mm}{\multirow{4}{*}{\rotatebox[origin=c]{90}{Diversity}}} &Vocabulary Size &0.0684 &0.07 &0.0884 &-0.1686 &0.0822 &0.0046 &-0.1408 &0.0483 &0.0072 \\
&DelFy (Source) &-0.076 &0.0856 &0.0011 &-0.0138 &0.0186 &0.0107 &-0.0053 &0.0071 &0.0144 \\
&DelFy (Target) &-0.0072 &-0.0522 &-0.0932 &-0.0606 &0.0674 &-0.1436 &-0.1091 &-0.0656 &0.1121 \\
&L2 Distance &-0.1665 &0.0208 &0.0411 &0.0381 &-0.0834 &0.0388 &-0.0736 &0.1915 &-0.104 \\
\cmidrule{2-11}
\parbox[t]{1mm}{\multirow{7}{*}{\rotatebox[origin=c]{90}{Informativeness}}} &Avg Token Entropy &-0.2114 &0.0027 &-0.0485 &0.0555 &0.1523 &0.0463 &-0.1427 &-0.1159 &0.0715 \\
&Avg Token Prob &0.1702 &-0.0128 &0.0268 &-0.0735 &-0.1593 &-0.0466 &0.1227 &0.0898 &-0.0632 \\
&Beam Search Weighted Avg &-0.1629 &0.0531 &-0.0455 &0.0716 &0.1568 &0.043 &-0.0345 &-0.0595 &0.0719 \\
&Beam Search Ratio &-0.2474 &0.1116 &-0.015 &-0.0458 &-0.0229 &0.0065 &-0.0373 &-0.0842 &0.1127 \\
&BALD &0.0314 &0.0262 &0.0689 &0.1 &-0.1572 &0.1089 &-0.1666 &0.0847 &0.0329 \\
&Dropout KL Div &0.0243 &0.0453 &0.114 &0.109 &-0.2244 &0.2104 &0.0896 &-0.2134 &-0.0961 \\
&Dropout Lexical Similarity &0.0542 &-0.2409 &-0.0865 &0.0285 &0.0972 &-0.1158 &0.1572 &0.0695 &-0.016 \\
\cmidrule{2-11}
& $R^2$ & 0.117 & 0.145 & 0.029 & 0.106 & 0.155 & 0.098 & 0.117 & 0.197 & 0.065 \\
\bottomrule
\end{tabular}
}
\caption{Spearman correlation between AL metrics and model performance using 500 samples for training (Test Set ChrF), * displayed for correlations significantly different from zero ($\alpha=0.05$ with Bonferroni correction)}\label{tab: }
\end{table*}

\begin{table*}[!htp]\centering
\resizebox{\textwidth}{!}{ 
\begin{tabular}{lrrrr}\toprule
& &\% Vocab Covered in Test &\% Vocab Jaccard Similarity to Test \\\midrule
\multirow{3}{*}{Flan-T5} &Eng-Afr &0.1021 &0.1051 \\
&Eng-Ger &0.0178 &-0.0026 \\
&Eng-Fil &0.2322 &0.2287 \\
\multirow{3}{*}{Llama 3.1} &Eng-Afr &-0.0946 &0.031 \\
&Eng-Ger &0.0899 &0.0537 \\
&Eng-Fil &0.0541 &0.0209 \\
\multirow{3}{*}{Gemma 2} &Eng-Afr &0.1814 &0.2878* \\
&Eng-Ger &0.0899 &0.0421 \\
&Eng-Fil &0.046 &0.0639 \\
\bottomrule
\end{tabular}
}
\caption{Spearman correlation between model test set performance and the similarity between the training and test set, using 500 samples for training * added if the correlation is significantly different from zero}\label{tab: }
\end{table*}

\begin{figure*}[h!]
    \centering
    \def\subfigwidth{0.32\textwidth}
    
    \begin{subfigure}{\subfigwidth}
        \includegraphics[width=\textwidth]{images/al_difficulty_probe_gemma_afr.png}
        \caption{Gemma, Eng-Afr}
    \end{subfigure}
    \hfill
    \begin{subfigure}{\subfigwidth}
        \includegraphics[width=\textwidth]{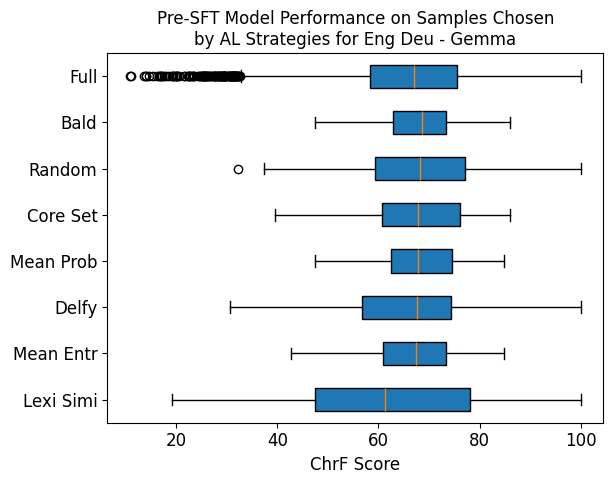}
        \caption{Gemma, Eng-Ger}
    \end{subfigure}
    \hfill
    \begin{subfigure}{\subfigwidth}
        \includegraphics[width=\textwidth]{images/al_difficulty_probe_gemma_fil.png}
        \caption{Gemma, Eng-Fil}
    \end{subfigure}
    
    \vspace{1em} 

    \begin{subfigure}{\subfigwidth}
        \includegraphics[width=\textwidth]{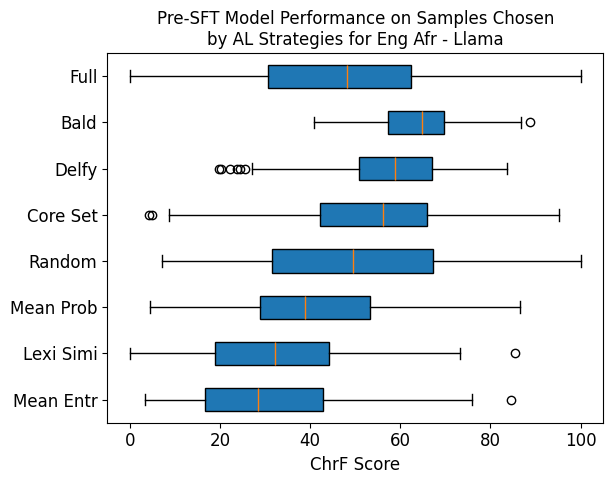}
        \caption{Llama, Eng-Afr}
    \end{subfigure}
    \hfill
    \begin{subfigure}{\subfigwidth}
        \includegraphics[width=\textwidth]{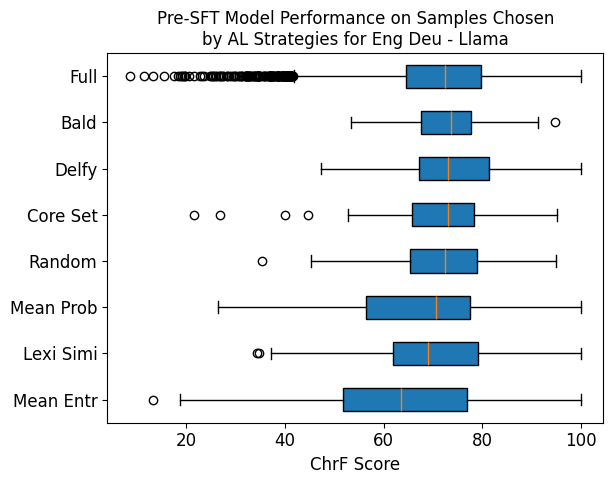}
        \caption{Llama, Eng-Ger}
    \end{subfigure}
    \hfill
    \begin{subfigure}{\subfigwidth}
        \includegraphics[width=\textwidth]{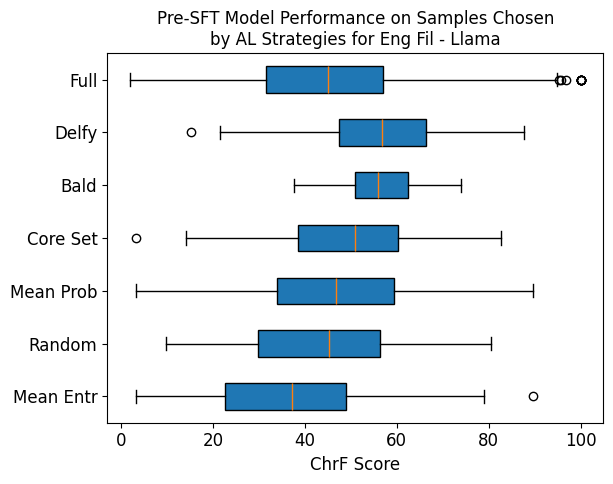}
        \caption{Llama, Eng-Fil}
    \end{subfigure}

    \vspace{1em}
    
    \begin{subfigure}{\subfigwidth}
        \includegraphics[width=\textwidth]{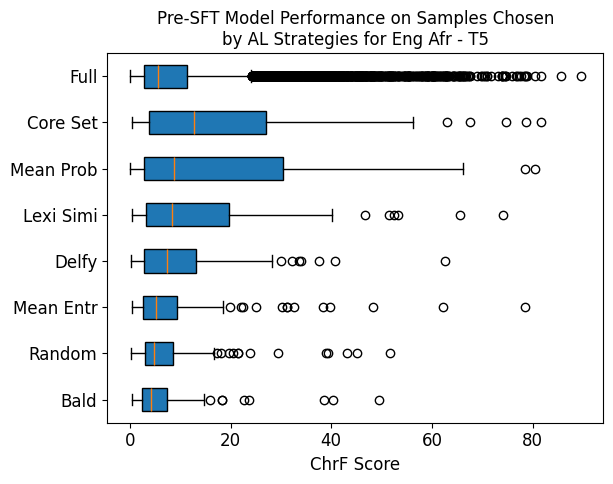}
        \caption{T5, Eng-Afr}
    \end{subfigure}
    \hfill
    \begin{subfigure}{\subfigwidth}
        \includegraphics[width=\textwidth]{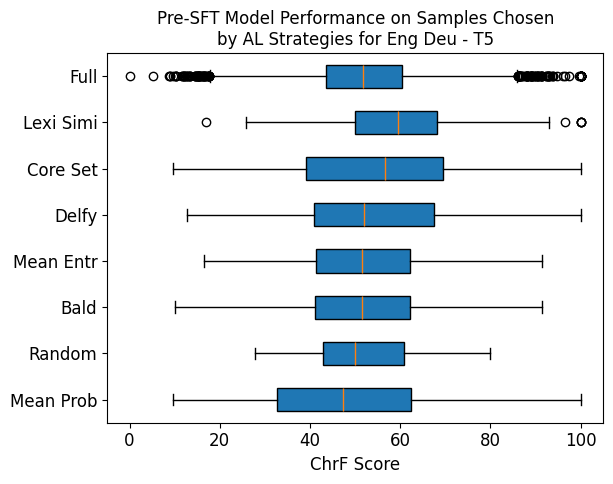}
        \caption{T5, Eng-Ger}
    \end{subfigure}
    \hfill
    \begin{subfigure}{\subfigwidth}
        \includegraphics[width=\textwidth]{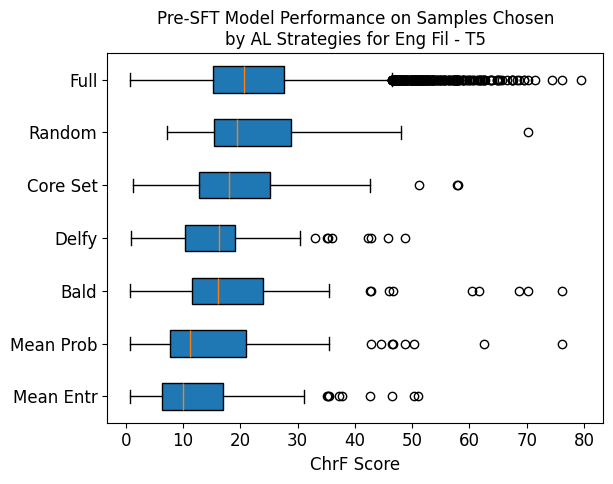}
        \caption{T5, Eng-Fil}
    \end{subfigure}

    \caption{Performance of model pre-SFT on candidates chosen by various strategies across three models and three tasks}
    \label{fig:grid_example}
\end{figure*}

\begin{figure*}[!htb]
\centering
  \includegraphics[width=\textwidth]{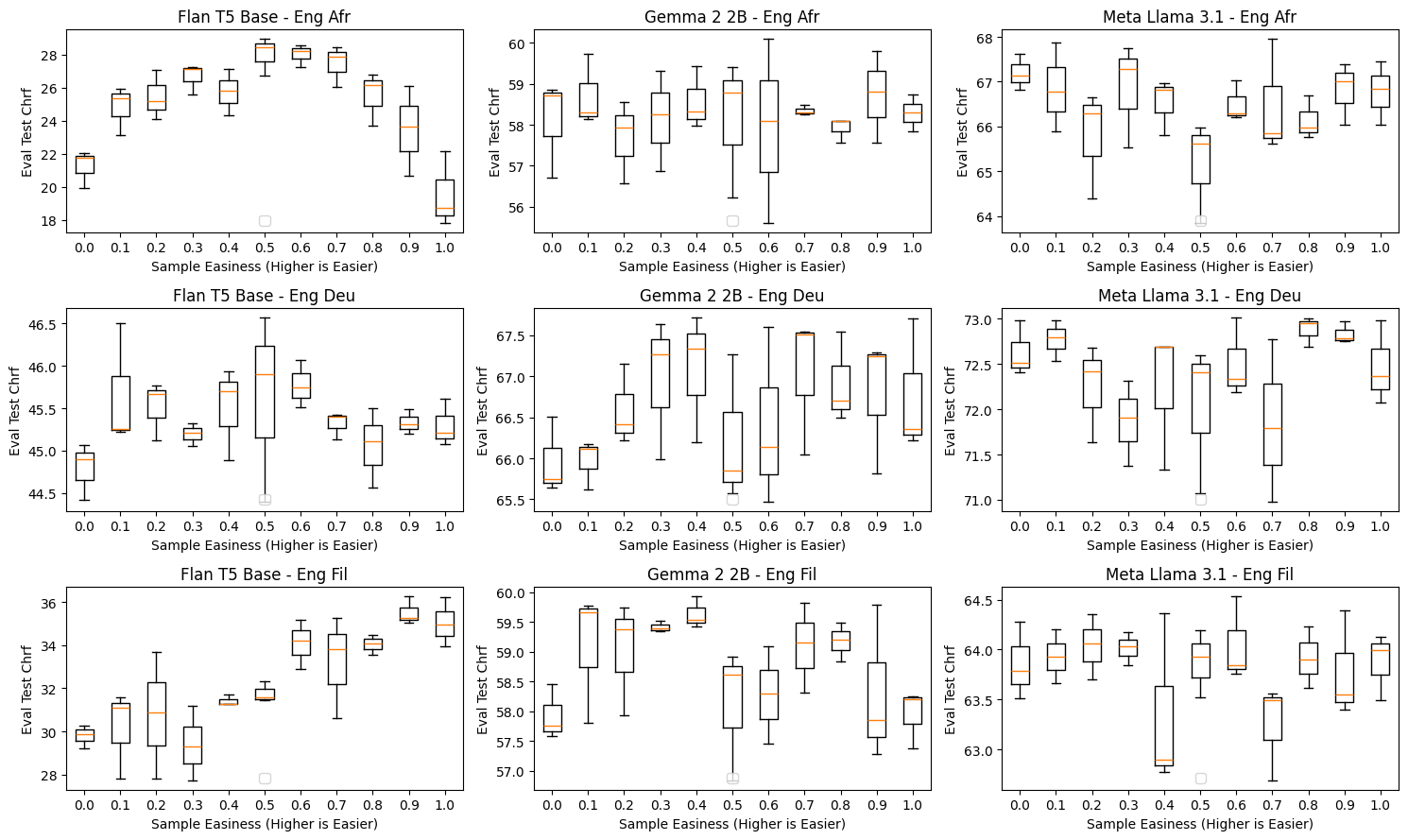}
  \caption{Test performance when fine-tuning models on unlabeled data with varying degrees of difficulty (measured using pre-SFT model performance) using 2000 unlabeled samples \label{Figure:difficulty_ablation_2000}}
\end{figure*}

\section{Case Study Results}

\begin{table*}
\centering
\scriptsize
\begin{tabular}{p{0.75cm} p{0.75cm} p{6.75cm} p{5.75cm}}
\toprule
\textbf{Set} & \textbf{Step} & \textbf{Input} & \textbf{Target} \\
\midrule
Train & 91 & A \textbf{\textcolor{green}{prayer}} for our beloved nation. & Isang \textbf{\textcolor{green}{panalangin}} sa aming mahal na nasyon. \\
\midrule
\textbf{Set} & \textbf{Step} & \textbf{Input} & \textbf{Prediction} \\
\midrule
Test & 94 & The tenth named storm of the Atlantic Hurricane season, Subtropical Storm Jerry, formed in the Atlantic Ocean today. & Ang lalake ang huling named na na na named na ang \textbf{\textcolor{red}{panalangin}}..., ang \textbf{\textcolor{red}{panalangin}} sa Atlantic ng Amerika... \\
\cmidrule{2-4}
 & 95 & The number of people present was so large that it was not possible for everybody to gain access to the funeral in St. Peter's Square. & Ang mga tao na ito ay hindi posible para sa lahat ng tao ang \textbf{\textcolor{red}{panalangin}} sa St. Peter's Square. \\ 
\cmidrule{3-4}
 & & Prime Minister Stephen Harper has agreed to send the government's 'Clean Air Act'... for review, before its second reading, after Tuesday's 25 minute meeting with NDP leader Jack Layton at the PMO. & Stephen Harper ay nag-iisa ang \textbf{\textcolor{red}{panalangin}} sa ang lahat ng mga tao para sa \textbf{\textcolor{red}{panalangin}}... \\ \cmidrule{2-4}
 & 97 & The final match of the series will take place at Ellis Park in Johannesburg next week, when the Springboks play Australia. & Ang palangin sa Ellis Park sa Johannesburg, ang \textbf{\textcolor{red}{panalangin}} sa Australia ng mga tao ng mga tao ng Australia.\\
\bottomrule
\end{tabular}
\caption{\label{Table:wrong_translation} Models incorrectly generate the word \textit{panalangin} across various samples after being fine-tuned on an example with the word (\textbf{\textcolor{red}{Red}} indicates wrong usage of the word)}
\end{table*}

\end{document}